\documentclass{article}

\usepackage{microtype}
\usepackage{graphicx}
\usepackage{booktabs} 

\usepackage{hyperref}
\usepackage{subcaption}

\usepackage[accepted]{icml2024}

\usepackage{threeparttable}
\usepackage{multirow}
\usepackage{paralist}

\usepackage{amsmath}
\usepackage{amssymb}
\usepackage{mathtools}
\usepackage{amsthm}

\usepackage[capitalize,noabbrev]{cleveref}

\theoremstyle{plain}

\theoremstyle{definition}

\theoremstyle{remark}

\usepackage[textsize=tiny]{todonotes}
\usepackage{wrapfig}

\begin{document}

\twocolumn[
\icmltitle{Mitigating Bias in Dataset Distillation}



\icmlsetsymbol{equal}{*}

\begin{icmlauthorlist}
\icmlauthor{Justin Cui}{UCLA}
\icmlauthor{Ruochen Wang}{UCLA}
\icmlauthor{Yuanhao Xiong}{UCLA}
\icmlauthor{Cho-Jui Hsieh}{UCLA}
\end{icmlauthorlist}

\icmlaffiliation{UCLA}{Department of Computer Science, University of California, Los Angeles}
\icmlcorrespondingauthor{Justin Cui}{justincui@ucla.edu}
\icmlcorrespondingauthor{Cho-Jui Hsieh}{chohsieh@cs.ucla.edu}

\icmlkeywords{Dataset Condensation, Dataset Distillation, Machine Learning, ICML}

\vskip 0.3in
]



\printAffiliationsAndNotice{}  

\begin{abstract}
  Dataset Distillation has emerged as a technique for compressing large datasets into smaller synthetic counterparts,  facilitating downstream training tasks. In this paper, we study the impact of bias inside the original dataset on the performance of dataset distillation. With a comprehensive empirical evaluation on canonical datasets with color, corruption and background biases, we found that color and background biases in the original dataset will be amplified through the distillation process, resulting in a notable decline in the performance of models trained on the distilled dataset, while corruption bias is suppressed through the distillation process. 
  To reduce bias amplification in dataset distillation, 
we introduce a simple yet highly effective approach based on a sample reweighting scheme utilizing kernel density estimation.
Empirical results on multiple real-world and synthetic datasets demonstrate the effectiveness of the proposed method. 
  Notably, on CMNIST with 5\% bias-conflict ratio and IPC 50, our method achieves 91.5\% test accuracy compared to 23.8\% from vanilla DM, boosting the performance by 67.7\%, whereas applying state-of-the-art debiasing method on the same dataset only achieves 53.7\% accuracy.
   Our findings highlight the importance of addressing biases in dataset distillation and provide a promising avenue to address bias amplification in the process.
\end{abstract}

\section{Introduction}
\label{sec.intro}
Dataset plays a central role in machine learning model performances. 
The rapidly growing size of contemporary datasets not only poses challenges to data storage and preprocessing, but also makes it increasingly expensive to train machine learning models and design new methods, such as architectures, hyperparameters, and loss functions.
As a result, dataset distillation (also known as dataset condensation~\cite{wang2018dataset}) emerges as a promising direction for solving this issue.
Dataset Distillation aims at compressing the original large-scale dataset into a small subset of information-rich examples, enabling models trained on them to achieve competitive performance compared with training on the whole dataset.
The distilled dataset with a significantly reduced size can therefore be used to accelerate model training and reduce data storage.
Dataset distillation has been shown to benefit a wide range of machine learning tasks, such as Neural Architecture Search~\cite{wang2021rethinking}, Federated Learning~\cite{xiong2023feddm}, Continual Learning~\cite{yang2023efficient,gu2023summarizing}, Graph Compression~\cite{jin2021graph,feng2023fair}, and Multimodality~\cite{wu2023multimodal}.

Recent dataset distillation methods mainly focus on performance enhancements on standard datasets~\cite{loo2023dataset,zhao2023improved,du2023minimizing,cui2023scaling}. However, this emphasis often overlooks dataset bias, a critical issue in machine learning. Dataset bias~\cite{tommasi2017deeper}  occurs when collected data unintentionally reflects existing biases, leading to skewed predictions and potential ethical concerns.
Despite extensive research on bias detection and mitigation strategies in recent years~\cite{sagawa2019distributionally,li2019repair,nam2020learning,lee2021learning,hwangselecmix}, the impact of dataset bias on data distillation remains unexplored. Given that a biased synthetic set can lead to inaccurate or unfair decisions, understanding the role of bias in dataset distillation and developing effective mitigation strategies is crucial.

This paper provides the first study of how biases in the original training set affect dataset distillation process. Specifically, we are interested in the following questions:
\textit{1). How does bias propagate from the original dataset to the distilled dataset?}
\textit{2). How do we mitigate biases present in the distilled dataset?}

To answer the first question, we assess existing dataset distillation algorithms across several benchmark datasets. Our findings reveal that the distillation process is significantly influenced by the type of bias, with color and background biases being amplified and noise bias suppressed. 
For datasets exhibiting amplified biases, we found that even state-of-the-art de-biasing training methods are insufficient in restoring the original performance. This highlights the urgency of developing a bias-mitigating dataset distillation algorithm.

To counter bias in distilled datasets, we propose a simple yet effective debiasing algorithm based on data point reweighting. Leveraging the insight that biased data points cluster in the model's embedding space, we down-weight samples within such clusters using kernel density estimation. This re-weighting rebalances the significance of biased and unbiased samples, ameliorating biases in the distillation process. 
\begin{figure}
    \centering
\includegraphics[width=1.0\linewidth]{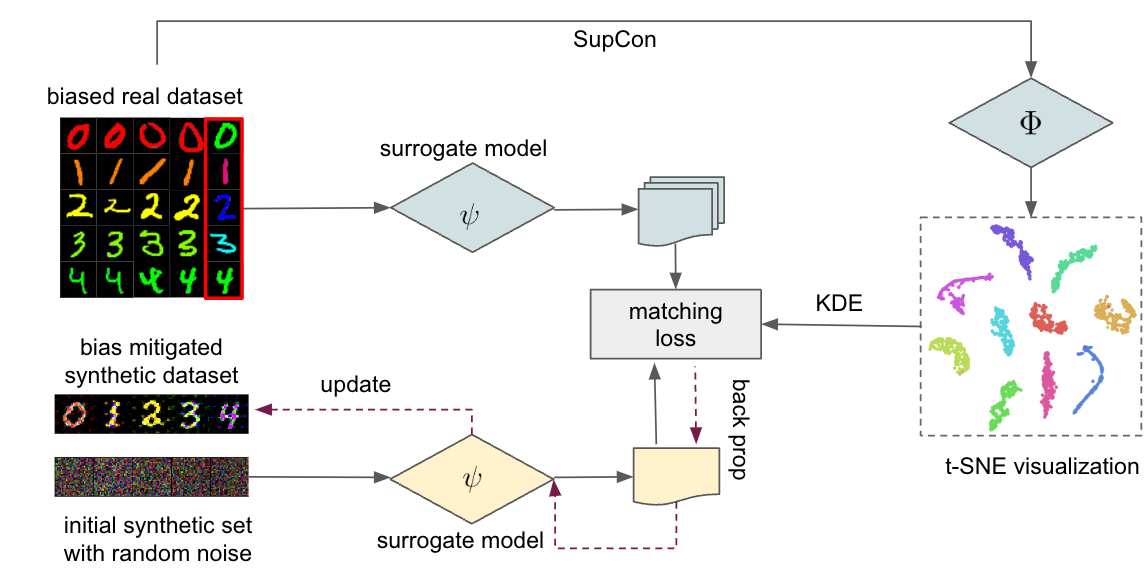}
    \caption{Workflow of our method that utilizes Supervised Contrastive model and Kernel Density Estimation to mitigate bias in the dataset distillation process.}
  \label{fig:overall_drawing}
  \vspace{-0.15in}
\end{figure}
Empirical results on diverse bias-injected datasets 
demonstrate that the proposed reweighting scheme significantly reduces bias in the distilled datasets. 
For example, on Colored MNIST with a 5\% bias in conflicting samples and 50 images per class, the original Distribution Matching (DM) method leads to a biased synthetic set. A model trained on such a distilled set achieves only 23.8\% accuracy. In contrast, our reweighting method produces a more balanced dataset, resulting in 91.5\% accuracy, representing a 67.7\% gain over DM. In summary, our contributions are: 
\begin{enumerate}
    \item We provide the first study on the impact of biases in dataset distillation process.
    \item We propose a simple yet effective re-weighting scheme to mitigate biases in two canonical types of dataset distillation methods.
    \item Through extensive experiments and ablation studies on both real-world and synthesized datasets, we demonstrate the effectiveness of our method.
\end{enumerate}

\section{Related Work}
\subsection{Dataset Distillation}
\label{sec.distillation}
Dataset Distillation (DD) or Condensation (DC) \cite{wang2018dataset,li2022awesome} aims to compress a large dataset into a small but informative one to achieve competitive performances compared to the whole dataset. Naturally, it’s framed as a bi-level optimization problem using surrogate objectives such as model gradients \cite{zhaodsa, feng2023embarrassingly, du2024sequential} or training trajectories \citep{cazenavette2022dataset, du2023minimizing, cui2023scaling, guo2023towards}. The inner loop of the bi-level optimization usually optimizes a surrogate model using the distilled dataset and the outer loop tries to optimize the resulting distilled dataset itself.

Due to the heavy computation involved in the bi-level optimization process, one line of work \citep{zhaodm, wang2022cafe, zhao2023improved, sajedi2023datadam, zhang2024m3d} tries to directly match the latent feature space. While for other works \citep{zhou2022dataset, loo2022efficient}, they draw inspiration from Kernel Ridge Regression (KRR) \citep{nguyen2020dataset} and try to approximate the surrogate model using kernel methods to tackle the high computation cost. Orthogonally, some methods \citep{kim2022dataset, deng2022remember, wei2024sparse, shin2024frequency, liu2023few} propose to compress the dataset into even more compact spaces through parametrization. In order to boost the applicability of the resulting dataset, recent works \citep{yin2024squeeze, sun2023diversity} try to scale DD to large scale datasets such as ImageNet-1K using different approaches such as a 3-stage process or an efficient paradigm that enables both diversity and realism. Pre-trained generative models are also used in recent works \citep{cazenavette2023generalizing, zhang2023dataset, wang2023dim} to enhance the distilled datasets.

Because of the highly compressed dataset size and competitive performance, the distilled dataset can be used for many downstream tasks such as Continual Learning \citep{yang2023efficient, gu2023summarizing}, Federated Learning \citep{xiong2023feddm, huang2023federated}, Graph Neural Networks \citep{zhang2024navigating, liu2024graph}, Neural Architecture Search \citep{such2020generative, medvedev2021learning}, etc. Since most of the parameterization methods can be used as an add-on module to non-parameterization methods, we focus on non-parameterization methods to study the effect of bias in this paper.

\subsection{Dataset and Model Bias}
Deep Neural Networks (DNNs) have achieved remarkable performance in discovering correlations present within datasets. However, when applied to datasets where simple and spurious correlations coexist with complex and intrinsic correlations, DNNs may inadvertently lean towards the shortcuts. The generalization ability of DNNs will be greatly hindered when these spurious relations are learned instead of the intrinsic ones.  Following previous works~\cite{hwangselecmix}, we refer to samples strongly correlates with bias feature as bias-aligned and samples that don't align with bias features as bias-conflicting. The most commonly studied bias types include color, background, noise, texture etc~\cite{nam2020learning,lee2021learning,hwangselecmix}. 
In this paper, we study how the bias in the original data affects the small synthetic set through the dataset distillation process and try to mitigate this phenomena. The evaluations are done by testing whether models trained on synthetic datasets can effectively generalize to unbiased test datasets.
\subsection{De-biasing Methods}
While prior research has explored de-biasing methods on entire datasets, our work is the first, to the best of our knowledge, to investigate the impact of biases on dataset distillation results in this emerging field. Below, we discuss several state-of-the-art de-biasing methods designed to train de-biased models on entire datasets.

LfF~\cite{nam2020learning} debiases by training a biased and debiased model together. It uses generalized cross entropy (GCE)~\cite{zhang2018generalized} loss to amplify bias and computes a difficulty score for debiased model loss weighting. DFA~\cite{lee2021learning} disentangles bias and intrinsic attributes, creating unbiased samples by swapping bias embeddings among training samples to train a debiased model. SelectMix~\cite{hwangselecmix} shares the same idea of creating more unbiased samples by using mixup augmentation on contradicting pairs selected by a biased model.

Note that, although some prior de-biasing methods involve reweighting~\cite{sagawa2019distributionally,li2019repair,nam2020learning}, our approach differs for the following reasons: 1) Previous methods~\cite{li2019repair,nam2020learning} integrate the reweighting scheme into the final de-biased model's training, either as a loss or using an auxiliary model, optimizing it alongside the de-biased model. This process does not apply to dataset distillation methods since a well-trained model is not required in the process (e.g., DM uses randomly initialized models, and MTT only matches part of the training trajectories). 2) Methods like~\cite{sagawa2019distributionally} require explicit bias supervision, which may be challenging or infeasible, while our method has no such requirement.

\section{The Impact of Bias in Dataset Distillation}
\label{lab:bias_effect}

In this section, we conduct comprehensive experiments to answer the following question: {\it how does a biased training set influence the distilled data? Will the bias be amplified or suppressed through the distillation process? } 

In line with prior studies~\cite{nam2020learning,lee2021learning,hwangselecmix}, we explore three datasets: Colored MNIST (CMNIST), Background Fashion-MNIST (BG FMNIST), and Corrupted CIFAR-10. CMNIST introduces a color bias, causing classes to share specific colors, potentially confusing training. BG FMNIST combines MiniPlaces (Zhou et al., 2017) with Fashion-MNIST, creating background biases, e.g. T-shirts in bamboo forests. Corrupted CIFAR-10 introduces perturbations and image distortions like Gaussian noise, blur, brightness, contrast changes, and occlusions. See~\cref{fig:example_images} for examples and~\cref{sec.datasets} for detailed descriptions.

\begin{figure*}
    \centering
    \includegraphics[width=0.33\textwidth]{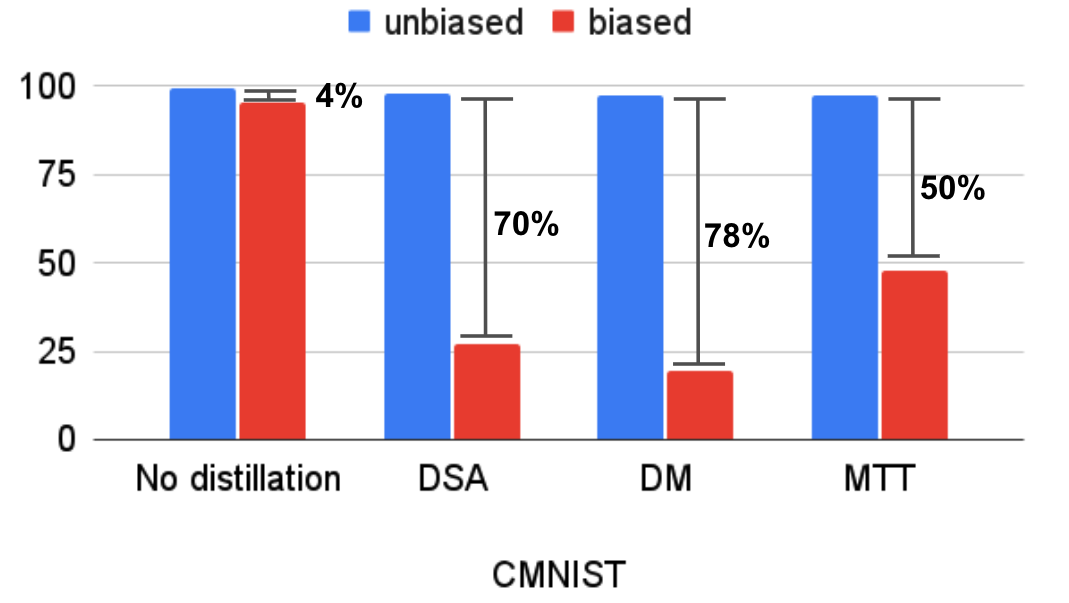}
    \includegraphics[width=0.33\textwidth]{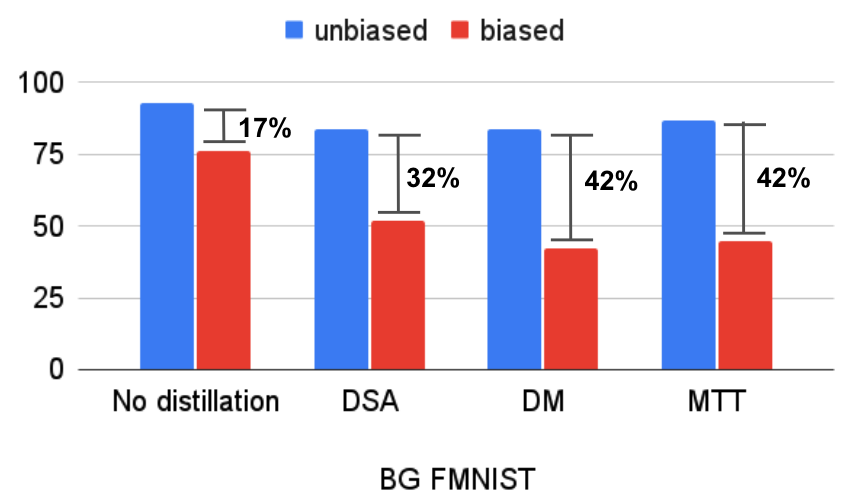}
    \includegraphics[width=0.33\textwidth]{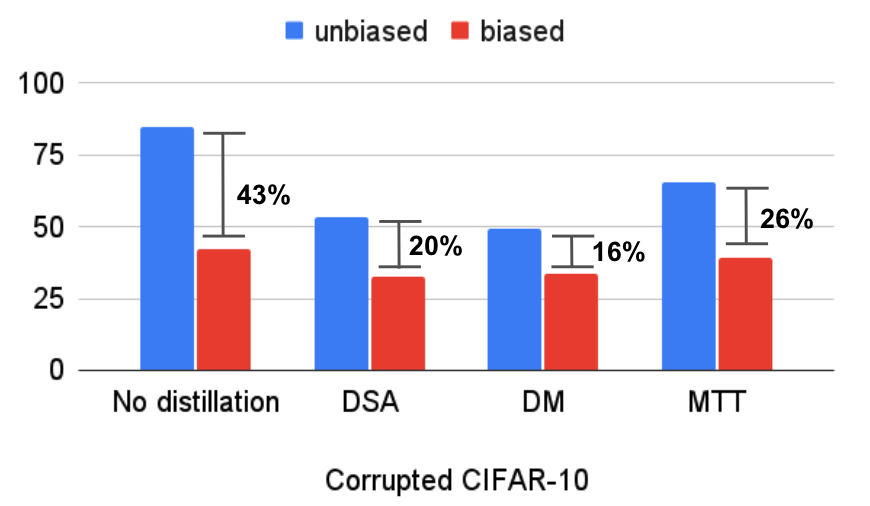}
    \caption{The left most 2 bars indicate the model performance on full dataset with no distillation. For DSA/DM/MTT, the blue bar shows the model performance on the unbiased dataset and the red bar shows the performance of the corresponding dataset distillation method on that biased dataset with 5\% bias-conflicting samples. The distillation performances are measured under IPC 10.}
    \label{fig:amplified}
\end{figure*}

For each dataset, we use $D$ to denote the unbiased set (e.g., randomly distributed colors in CMNIST), while $D_b$ represents the bias-injected dataset. In $D_b$, 95\% of the samples are aligned with the bias, meaning, for example, 95\% of the digit '0' may be red, while the remaining 5\% of the digit '0' possess random colors. Let $\mathcal{F}$ be a dataset distillation algorithm that maps the original dataset into a distilled synthetic set, and let $\mathcal{M}$ denote the model training procedure that maps a dataset to a model. By comparing $\mathcal{M}(\mathcal{F}(D))$ (model trained with distilled unbiased dataset) and $\mathcal{M}(\mathcal{F}(D_b))$ (model trained with distilled biased dataset), we evaluate these two models' performance on unbiased test samples and compute the following measurement to reveal how the bias in the source dataset 
 degrades the performance of the model trained on distilled synthetic samples:
\begin{equation*}
    \text{Acc}(\mathcal{M}(\mathcal{F}(D))) - \text{Acc}(\mathcal{M}(\mathcal{F}(D_b))). 
\end{equation*}
We conduct experiments on three representitive dataset distillation algorithms: DSA~\cite{zhaodsa}, DM~\cite{zhaodm} and MTT~\cite{cazenavette2022dataset} as lots of SOTA methods can be used as an add-on module to these methods~\cite{wang2022cafe,kim2022dataset, liu2022dataset,lee2022dataset}. Additionally, we include a baseline approach without dataset distillation, which is equivalent to the case when $\mathcal{F}$ represents an identity transformation.  The visualization of bias impacts in dataset distillation can be found in~\cref{fig:amplified}.

For CMNIST, the results presented on the left panel of~\cref{fig:amplified} reveal a strong {\bf bias amplification} effect -- while the bias injected into the original dataset leads to a mere 4\% performance drop in regular training, it results in over 50\% performance decline when employing any of the three dataset distillation methods. This observation can be attributed to the following factors. Since the original set comprises 95\% biased samples, with a selection of IPC 10, it is highly possible that all of the chosen images are biased. As 
 a result, the unbiased signal totally diminishes through the distillation process. Another critical factor is that since color is a discriminative feature that can be easily learned by neural networks, dataset distillation algorithms will focus on distilling this color feature into the synthetic images in order to achieve good performance, leading to bias amplification. Similar impacts are also seen on the BG FMNIST dataset. 

Interestingly, results from Corrupted CIFAR-10 show a reverse trend. The right panel of~\cref{fig:amplified} shows that the performance degradation is actually more substantial in the traditional training pipeline compared to the ones incorporating dataset distillation. 
This observation indicates the {\bf bias suppression} effect for corruption biases, and dataset distillation is helpful for mitigating bias in this setting. 
Since corruption biases include several different perturbation effects such as Gaussian noise and blurring, we assume that the distillation process naturally blends information from multiple images, and the resulting images are already blurred in nature where noisy effects tend to cancel out, so it is harder to capture the corruption bias. Sampled distilled images are visualized in~\cref{fig:example_images}. 

For datasets exhibiting amplified biases (CMNIST and BG FMNIST), we found that even state-of-the-art de-biasing training methods such as SelectMix~\cite{hwangselecmix} and DFA~\cite{lee2021learning} are not able to obtain an unbiased model from the biased synthetic set. See more details in~\cref{sec.ablation}. This finding also highlights the urgency of developing a bias-mitigating DC algorithm to obtain unbiased synthetic sets. 

\section{Ameliorate Bias in Dataset Distillation}

\subsection{Bias Amelioration Through Re-weighting}
\label{sec.dm_ours}
In dataset distillation methods such as DM~\cite{zhaodm, zhao2023improved}, the objective is to match the embeddings generated by the synthetic dataset ($\mathcal{S}$) with the ones generated by the real images ($\mathcal{T}$). The objective function is formulated as below
\begin{equation}
\label{eq:dm_loss}
    \underset{\mathcal{S}}{\min} \ \ \underset{\omega\sim \Omega}{\underset{v
    \sim P_{v}}{\mathbb{E}}}\parallel \frac{1}{|\mathcal{T}|} \overset{|\mathcal{T}|}{\underset{i=1}{\sum}}\psi_{v}(\mathcal{A}(x_i, \omega)) - \frac{1}{|\mathcal{S}|} \overset{|\mathcal{S}|}{\underset{i=1}{\sum}}\psi_{v}(\mathcal{A}(s_j, \omega)) \parallel^2,
\end{equation}
where $\psi$ is usually a surrogate model that maps the data into an embedding space and $\mathcal{A}$ is a differentiable augmentation function. By solving \eqref{eq:dm_loss}, DM learns the synthetic set to match the mean embedding of the original set.  If the dataset is highly biased, the first term $\frac{1}{|\mathcal{T}|} \overset{|\mathcal{T}|}{\underset{i=1}{\sum}}\psi_{v}(\mathcal{A}(x_i, \omega))$ will be dominated by bias-aligned samples, thus causing distribution matching based methods to synthesize more bias-aligned images. 

To mitigate bias, we propose to compute a weighted sum of the real image embeddings instead of simply using the mean of all data points. For data points that exhibit a strong correlation with the spurious (bias) feature, they should be assigned lower weights. Conversely, data points that have limited association with the bias feature should be afforded higher weights. This adjustment ensures that the distillation process effectively captures the intrinsic features. Let $W(\mathcal{T})=[w_0,w_1,...,w_n]$ with $n$ equals $|\mathcal{T}|$ in~\cref{eq:dm_loss} be the normalized weight of each training sample, the weighted loss can be written as
\begin{equation}
\label{eq:dm_loss_ours_generic}
    \underset{\mathcal{S}}{\min} \ \ \underset{\omega\sim \Omega}{\underset{v
    \sim P_{v}}{\mathbb{E}}}\parallel W(\mathcal{T})\cdot\psi_{v}(\mathcal{A}(\mathcal{T}, \omega)) - \frac{1}{|\mathcal{S}|} \overset{|\mathcal{S}|}{\underset{i=1}{\sum}}\psi_{v}(\mathcal{A}(s_j, \omega)) \parallel^2,
\end{equation}
where $W(\mathcal{T})\cdot\psi_{v}(\mathcal{A}(\mathcal{T}, \omega))$ is the re-weighted embeddings where the bias feature has been balanced.

\subsection{Bias Estimation Using Kernel Density Estimation}
\label{sec.kde}
The key problem of this reweighting scheme is how to compute $W(X)$ which is unknown. In order to compute it, we propose to use Kernel Density Estimation (KDE), which is a non-parametric technique used to estimate the probability density function (PDF) of a random variable based on observed data points. Mathematically, given a set of n data points $x_1,x_2,...,x_n$, the kernel density estimate  $\hat{f}(x)$ at any point x is given by:
\begin{equation}
\label{eq:kde}
\hat{f}(x) = \frac{1}{n} \sum_{i=1}^{n} K\left(\|\Phi(x) - \Phi(x_i)\|\right), 
\end{equation}
where $K(\cdot)$ is the kernel function that determines the shape and width of the kernel placed on each data point. The most commonly used kernel function is the Gaussian kernel: $ K(u) = \frac{1}{\sigma\sqrt{2\pi}} e^{-\frac{1}{2}\cdot(\frac{u}{\sigma})^2}$. $\Phi$ is a function that 
maps the data points into an embedding space for distance computation. By summing the contributions of the kernel functions centered at each data point, KDE provides an estimate of the PDF at any given point $x$. The estimate $\hat{f}(x)$ represents the density of the underlying distribution at that point. 
Since a biased dataset is usually dominated by bias-aligned samples which should be given a lower weight, we propose to use the normalized inverse of the kernel density function~$\mathbb{N}(\frac{1}{\hat{f}(x)})$ as the new weights,
and $\mathbb{N}$ is a normalization function such as softmax so that $\underset{i=1}{\overset{|\mathcal{T}|}{\sum}}\mathbb{N}(\frac{1}{\hat{f}(x_i)}) =1$. Eventually we have $W(\mathcal{T})=[\mathbb{N}(\frac{1}{\hat{f}(x_1)}),..\mathbb{N}(\frac{1}{\hat{f}(x_n)})]$.

\subsection{Distance Computation in KDE}
The aforementioned KDE reweighting scheme requires a feature mapping $\Phi(\cdot)$ to map images from raw pixel space to a more meaningful hidden space. Moreover, an optimal mapping for $\Phi(\cdot)$ would be one that transforms images into biased features, enabling KDE to accurately represent the density of bias. Consequently, our reweighting scheme can effectively mitigate bias. To obtain bias features without external knowledge, 
DFA~\cite{lee2021learning} and LfF~\cite{ nam2020learning} both utilize the generalized cross entropy (GCE) loss to train an auxiliary bias model. However, neither of them directly work with latent spaces that can be utilized in KDE. Following the recent SOTA de-biasing method SelecMix~\cite{hwangselecmix}, we utilize a supervised constrastive learning~\cite{chen2020simple, khosla2020supervised} model, which is trained with generalized supervised contrastive (GSC) loss to produce image embeddings that can be used to measure distances between data points. The model is proven to produce high quality similarity matrix regarding bias features~\cite{hwangselecmix}. See~\cref{fig:overall_drawing} for the visualization of our overall workflow.

\subsection{Apply to Other Dataset Distillation Methods}
In addition to DM and its follow up works such as DREAM~\cite{liu2023dream} and IDM~\cite{zhao2023improved}, our method can also be easily applied to other dataset distillation methods. Here we use DSA~\cite{zhaodsa,zhao2023dataset, kim2022dataset} as an example which synthesizes data by matching the gradients between surrogate models trained with real data and synthetic data. Generally, gradient matching based methods can be formulated as:
\begin{equation}
\label{eq:dsa_loss}
\underset{\mathcal{S}}{\min} \ \ D(\nabla_{\theta}\mathcal{L}_c^{\mathcal{T}}(\mathcal{A}(\mathcal{T}, 
\mathcal{\omega}^\mathcal{T}), \theta_{t}), \nabla_{\theta}\mathcal{L}_c^{\mathcal{S}}(\mathcal{A}(\mathcal{S},
\mathcal{\omega}^\mathcal{S}), \theta_{t})), 
\end{equation}
where $\mathcal{L}_c^{\mathcal{T}} = \frac{1}{|\mathcal{T}|}\sum_{x,y}\ell(\phi_{\theta_t}(\mathcal{A}(\mathcal{T},\mathcal{\omega}^\mathcal{T}))$. 
We denote the re-computed weights as $W(\mathcal{T})$, then the first term in \cref{eq:dsa_loss} can be replaced with $W(\mathcal{T})\cdot\nabla_{\theta}\mathcal{L}_c^{\mathcal{T}}(\mathcal{A}(\mathcal{T}, 
\mathcal{\omega}^\mathcal{T}), \theta_t)$. We demonstrate that the proposed method works well with both DSA and DM in the following sections. 

\section{Experiments}
\subsection{Evaluation Datasets}
\label{sec.datasets}
Following previous works~\cite{lee2021learning, hwangselecmix}, we first evaluate our method on 3 canonical biased datasets including two synthesized datasets Colored MNIST~(CMNIST) and Corrupted CIFAR-10 and one real-world dataset BFFHQ. CMNIST originates from MNIST~\cite{deng2012mnist} dataset with added color correlation such as most digit 0 being red and digit 4 being green. Corrupted CIFAR10 applies different corruptions to the images in CIFAR-10 so that one class are associated with one type of corruption such as GaussianNoise or MotionBlur. BFFHQ~\cite{lee2021learning} is constructed from FFHQ~\cite{karras2019style} which contains human face images paired with their corresponding facial attribute annotations. Age is selected as the label and gender 
 is selected as the biased feature. We also contribute one novel dataset Background Fashion-MNIST~(BG FMNIST) to evaluate background biases which is constructed by using Fashion-MNIST~\cite{xiao2017fashionmnist} as foregrounds and MiniPlaces~\cite{zhou2017places} as backgrounds. Similar to SelecMix~\cite{hwangselecmix} and DFA~\cite{lee2021learning}, we evaluate our methods under 3 settings with 1\%,2\%,5\% bias conflicting samples\footnote{Bias-conflicting samples are samples in a class that have different bias properties from the majority of the samples in that class. For example, a yellow 0 is a bias conflicting sample when the majority of 0s are red. } for CMNIST, Corrupted CIFAR-10 and BG FMNIST and 0.5\% bias conflicting ratios for BFFHQ\footnote{BFFHQ only comes with 0.5\% bias conflicting ratio~\cite{lee2021learning}.}. More detailed description regarding datasets can be found in ~\cref{appendix.sec.datasets}.

\subsection{Experimental Results}
\begin{table*}
  \centering
  \caption{Test accuracy of baseline methods and our methods on distilled datasets with varying bias-conflict ratios.}
  \label{table:compare_to_baseline_synthetic}
  \vskip 0.1in
  \resizebox{1.0\textwidth}{!}{
  \begin{threeparttable}
  \begin{tabular}{ccccccccccc}
    \toprule
    Dataset & Method & \multicolumn{3}{c}{Bias-conflict Ratio (1.0\%)} &\multicolumn{3}{c}{Bias-conflict Ratio (2.0\%)} & \multicolumn{3}{c}{Bias-conflict Ratio (5.0\%)} \\
    &  & 1 & 10 & 50 & 1 & 10 & 50 & 1 & 10 & 50\\
    \midrule
    \multirow{8}{*}{CMNIST} 
    & Random & 22.6$\pm$1.0&13.5$\pm$0.3&16.0$\pm$0.1&22.8$\pm$1.0&16.2$\pm$0.1&20.7$\pm$0.1&19.9$\pm$0.5&17.4$\pm$0.4&27.2$\pm$0.2\\
    & MTT & 24.2$\pm$0.3&27.6$\pm$0.4&18.1$\pm$0.6&\textbf{28.3$\pm$0.3}&42.0$\pm$0.4&26.0$\pm$0.5&29.2$\pm$0.9&47.7$\pm$0.8&33.9$\pm$1.2\\
    & IDM & 
    20.1$\pm$0.9&18.8$\pm$1.2&17.6$\pm$1.6 &
    20.2$\pm$1.0&20.1$\pm$0.5&19.1$\pm$1.3 & 
    20.8$\pm$0.9&22.3$\pm$1.1&26.5$\pm$1.5\\
    & DREAM & 
    21.3$\pm$0.4& 15.9$\pm$1.2 & 17.9$\pm$0.8 & 26.2$\pm$0.5 & 17.1$\pm$1.3 & 17.5$\pm$1.5 & 23.7$\pm$1.2 & 30.8$\pm$1.5 & 50.4$\pm$1.0\\
    & DM & 25.4$\pm$0.1 & 18.6$\pm$0.2 & 22.6$\pm$0.5&24.8$\pm$0.4&18.5$\pm$0.6 & 23.6$\pm$0.8 & 25.3$\pm$0.3 & 19.6$\pm$0.9 & 23.8$\pm$1.3\\
    & DM+Ours & \textbf{28.0$\pm$0.5}&\underline{64.9$\pm$0.3}&\underline{75.4$\pm$1.1}&\underline{26.4$\pm$0.7}&\underline{50.6$\pm$1.2}&\underline{75.7$\pm$1.0}&\underline{32.2$\pm$1.0}&\underline{86.5$\pm$1.2}&\underline{91.5$\pm$0.9}\\
    & DSA & 26.1$\pm$0.3&16.5$\pm$0.2&14.5$\pm$0.2&25.2$\pm$0.3&16.8$\pm$0.3&30.9$\pm$0.4&25.9$\pm$0.5&27.3$\pm$0.4&68.5$\pm$1.2\\
    & DSA+Ours & \underline{27.9$\pm$0.4}&\textbf{76.7$\pm$1.1}&\textbf{81.4$\pm$0.8}&\underline{26.4$\pm$0.2}&\textbf{75.3$\pm$0.3}&\textbf{83.0$\pm$1.2}&\textbf{32.6$\pm$0.1}&\textbf{91.9$\pm$0.7}&\textbf{94.0$\pm$0.8}\\
    \midrule
    \multirow{8}{*}{BG FMNIST} 
    & Random &40.0$\pm$0.2 & 40.4$\pm$1.6 & 35.2$\pm$0.4 & 30.2$\pm$0.6 & 43.2$\pm$0.2 & 33.5$\pm$1.0&36.2$\pm$1.3&44.6$\pm$1.2&41.2$\pm$1.1\\
    &MTT&39.0$\pm$1.2 & 48.0$\pm$1.5 & 45.3$\pm$0.9 & 38.9$\pm$1.4 & 59.2$\pm$1.1 & 59.3$\pm$0.8 & \underline{48.1$\pm$1.4} & 45.2$\pm$1.3 & 62.3$\pm$0.8\\
    &IDM &
    40.7$\pm$1.0&42.3$\pm$0.9&38.4$\pm$1.2 &
    41.1$\pm$0.8&37.2$\pm$1.3&40.5$\pm$0.9 &
    43.4$\pm$0.4&46.6$\pm$0.8&42.0$\pm$1.2\\
    &DREAM&
    39.7$\pm$0.9&46.4$\pm$1.2&46.1$\pm$1.5&
    45.0$\pm$1.0&46.0$\pm$0.8&44.5$\pm$0.8&
    43.5$\pm$0.9&52.4$\pm$1.5&53.2$\pm$1.2\\
    & DM & 41.0$\pm$0.3 & 42.2$\pm$0.8 & 43.9$\pm$0.4&40.1$\pm$0.6 & 40.1$\pm$0.9&44.4$\pm$0.5& 41.7$\pm$0.5 & 42.0$\pm$1.2 & 44.6$\pm$0.9\\
    &DM+Ours & \textbf{44.6$\pm$0.5} & \underline{50.6$\pm$0.2} & \underline{57.2$\pm$0.6} & \textbf{51.4$\pm$0.7} & \underline{62.3$\pm$0.4} & \underline{63.0$\pm$1.0} & \textbf{49.4$\pm$0.2} & \underline{61.8$\pm$0.6} & \underline{65.0$\pm$0.8} \\
    & DSA & 43.4$\pm$0.4 & 45.8$\pm$0.5 & 40.7$\pm$0.9&43.7$\pm$0.5 & 47.6$\pm$0.3 & 48.4$\pm$0.8 & 44.7$\pm$0.6 & 52.8$\pm$0.5 & 59.3$\pm$0.6\\
    &DSA+Ours & \underline{44.4$\pm$0.6} & \textbf{57.0$\pm$1.0} & \textbf{58.3$\pm$0.8} & \underline{48.5$\pm$1.2} & \textbf{64.4$\pm$0.9} & \textbf{65.1$\pm$0.8} & 46.2$\pm$0.6 & \textbf{66.4$\pm$0.6} & \textbf{71.2$\pm$1.1} \\
    \midrule
    \multirow{8}{*}{Corrupted CIFAR-10} 
    & Random & 16.4$\pm$0.6&26.9$\pm$0.2&32.7$\pm$0.3&19.1$\pm$0.1&23.2$\pm$0.2&33.5$\pm$0.1&11.8$\pm$0.1&26.4$\pm$0.3&34.2$\pm$0.4\\
    & MTT & 23.5$\pm$0.4&25.4$\pm$1.5&33.3$\pm$0.5&24.1$\pm$0.3&\textbf{36.3$\pm$0.4}&35.7$\pm$0.2&24.2$\pm$0.8&\textbf{39.0$\pm$0.3}&\underline{39.5$\pm$0.4}\\
    & IDM & 
    \textbf{26.3$\pm$1.0}&30.2$\pm$0.4&36.6$\pm$0.8 &
    \textbf{26.2$\pm$0.8}&29.5$\pm$0.2&35.0$\pm$0.7 &
    \underline{26.0$\pm$0.8}&31.4$\pm$1.1&36.5$\pm$0.9 \\
    & DREAM &
    23.7$\pm$0.9&25.0$\pm$1.2&33.6$\pm$0.9 &
    \underline{25.9$\pm$0.8}&25.2$\pm$0.6&32.1$\pm$1.3 &
    25.6$\pm$0.9&24.5$\pm$0.8&33.7$\pm$1.2\\
    & DM & 25.1$\pm$0.4&\underline{32.9$\pm$0.3}&\underline{37.6$\pm$0.8}&25.0$\pm$0.1&32.9$\pm$0.1&\underline{37.7$\pm$0.2}&24.6$\pm$0.4&\underline{33.5$\pm$0.8}&38.7$\pm$0.4\\
    & DM+Ours & 24.2$\pm$1.2&\textbf{33.4$\pm$0.9}&\textbf{39.4$\pm$0.8}&25.3$\pm$0.5&\underline{34.2$\pm$0.5}&\textbf{39.7$\pm$0.4}&\textbf{26.6$\pm$0.5}&\underline{33.5$\pm$0.6}&\textbf{40.2$\pm$0.4}\\
    & DSA & 25.5$\pm$0.3&31.9$\pm$0.8&34.1$\pm$0.5&25.1$\pm$0.2&32.0$\pm$0.1&34.2$\pm$0.3&25.7$\pm$0.5&32.8$\pm$0.6&35.6$\pm$0.5\\
    & DSA+Ours & \underline{26.0$\pm$0.1}&32.6$\pm$0.8&35.0$\pm$0.6&25.2$\pm$0.8&33.2$\pm$0.2&35.8$\pm$0.6&\underline{26.0$\pm$0.3}&32.5$\pm$0.7&36.6$\pm$0.3\\
    \bottomrule
  \end{tabular}
   \begin{tablenotes}
    \footnotesize
    \item 1. \textbf{Results} in bold show the best performance. \underline{Results} with underline show the second best performance.
    \item 2. See~\cref{appendix.extension} for extending our method to DREAM~\cite{liu2023dream} and IDM~\cite{zhao2023improved} which achieves SOTA results based on DSA and DM.
    \item 3. See~\cref{appendix.extreme} for the results under extremely low bias-conflict ratio (0.5\%) where our method also achieves strong performances. 
    \end{tablenotes}
  \end{threeparttable}
  }
  \vskip -0.1in
\end{table*}
We evaluate the performance of 4 representative dataset distillation methods: DM~\cite{zhaodm}
, DSA~\cite{zhaodsa}, DREAM~\cite{liu2023dream} and IDM~\cite{zhao2023improved}. We then use the proposed algorithm to improve DM and DSA algorithms in the experiments. Note that our algorithm can also be applied to other existing distillation methods, as discussed in~\cref{appendix.extension}.
Also, see \cref{appendix.qa} for a qualitative analysis and~\cref{appendix.exp_setup} for specific hyper-parameter settings. 
\subsubsection{Performance boost on DM}
\label{sec.dm_boost}
We first investigate if the proposed method can mitigate biases in distribution matching-based methods such as DM. The evaluation results are shown in~\cref{table:compare_to_baseline_synthetic}. It can be seen that as the number of IPCs increases, the vanilla DM has little to no performance gain on bias amplifying datasets CMNIST and BG FMNIST. When IPC increases from 1 to 10 on CMNIST with 1, 2 and 5 percent bias-conflicting samples, there is even a  performance degradation. After reweighting the samples according to~\cref{eq:dm_loss_ours_generic}, we are able to mitigate the biases effectively. Under the settings of IPC 50, we are able to boost the performance from 22.6\% to 75.4\% on CMNIST dataset with 1\% bias conflicting samples. With 2\% and 5\% bias conflicting samples, the accuracy also increased from 23.6\% to 75.7\% and 23.8\% to 91.5\% respectively. The complete results can be found in \cref{table:compare_to_baseline_synthetic,table:real_world_dataset}. Similar performance boosts are also observed on BG FMNIST, e.g. with IPC 10, the performance gains are 8.4\%, 22.2\% and 19.8\% with 1, 2 and 5 percent bias conflicting samples. As shown in 
 previous works~\cite{lee2021learning,hwangselecmix}, de-biasing becomes more challenging on real-world dataset BFFHQ which only comes with 0.5\% bias conflict samples.  However, we still see an overall improvement of up to 6.2\%. On Corrupted CIFAR-10, we only observe a slight performance boost which aligns with the intuition described in~\cref{lab:bias_effect}.

\begin{table}
  \caption{Test accuracy on real-world dataset BFFHQ with varying IPCs. BFFHQ dataset only comes with 0.5\% bias-conflict samples which makes bias mitigation challenging. Our method is still able to boost the performance by up to 6.2\%.}
  \label{table:real_world_dataset}
  \vskip 0.1in
  \centering
  \resizebox{1.0\linewidth}{!}{
  \begin{threeparttable}
  \begin{tabular}{cccccccc}
    \toprule
    IPC & Random & MTT & DREAM & DM & DM+Ours & DSA & DSA+Ours \\
    \midrule
     1 & 50.8$\pm$1.5 & 52.5$\pm$1.2 & 54.0$\pm$1.2&58.3$\pm$0.3 & \textbf{64.5$\pm$1.5} & 61.1$\pm$1.0 & \underline{62.3$\pm$0.2}\\
     10& 50.2$\pm$0.8 & 58.6$\pm$1.3 & 61.2$\pm$1.9& 63.6$\pm$1.0 & \textbf{65.2 $\pm$1.7} & 62.6$\pm$1.6 & \underline{64.0$\pm$1.5}\\
     50 & 50.5 $\pm$1.2 & - & \underline{64.7$\pm$1.2}&63.9$\pm$1.2 & \textbf{65.6$\pm$ 0.9}& 58.7$\pm$1.4 &61.5$\pm$0.7 \\
    \bottomrule
  \end{tabular}
  \begin{tablenotes}
    \footnotesize
    \item [*] We aren't able to get IDM's performance because it's OOM on all settings for BFFHQ dataset.
    \end{tablenotes}
    \end{threeparttable}
  }
  \vskip -0.1in
\end{table}
\subsubsection{Performance boost on DSA}
Next, we investigate whether the proposed method can improve gradient matching based methods and choose DSA as the representative algorithm. Similar to DM, we also observe a strong performance boost on CMNIST. Under the settings of IPC 50, with 1, 2 and 5 percent bias conflicting samples, the performance increases from 14.5\% to 81.4\%, 30.9\% to 83.0\% and 68.5\% to 94.0\%. On BG FMNIST, the performances increase from 40.7\% to 58.3\%, 48.4\% to 65.1\% and 59.3\% to 71.2\% for 1, 2 and 5\% bias conflicting samples with IPC 50. On BFFHQ, we see a performance increase of up to 2.8\% even with the extremely low bias-conflict ratio. Complete results are shown in~\cref{table:compare_to_baseline_synthetic,table:real_world_dataset}.

\subsubsection{Compare to other SOTA methods}
\label{sec.compare_to_other}
The performance of 3 other representative methods, DREAM, IDM and MTT, are also included in~\cref{table:compare_to_baseline_synthetic,table:real_world_dataset}. 
We observe that MTT outperforms vanilla DM and DSA, achieving better results on CMNIST and BG FMNIST. For instance, on BG FMNIST with 5\% bias-conflicting samples and IPC 50, MTT achieves 62.3\% accuracy compared to DM's 44.6\% under the same settings. We think the reason is that MTT doesn't use real images during distillation phase but the trajectories from teacher models. Thus its performance is determined by both the teacher model trajectories (directly through trajectory matching) and biased original dataset (indirectly through teacher models trained using the biased original dataset). However, it also struggles with biases on general, whereas our proposed method is able to mitigate biases effectively on CMNIST, BG FMNIST and BFFHQ. DREAM and IDM, based on DSA and DM, tend to perform worse on biased datasets. This is partly due to their attempt to extract more information from the original dataset, potentially introducing additional biases. For instance, DREAM's use of Kmeans for selecting representative samples can lead to exclusion of unbiased samples in a biased dataset. See~\cref{appendix.extension} for results of applying our method to mitigate biases in DREAM and IDM.

 \subsection{Runtime Analysis}
 Our method has a similar runtime overhead to the state-of-the-art de-biasing method SelecMix~\cite{hwangselecmix}. It consists of two parts: 1) training a ResNet18~\cite{He_2016_CVPR} with a 128-dimensional MLP projection head, typically converging within 30 minutes which is around 5\% overhead on DSA which usually requires more than 10 hours to finish with IPC 50 on the hardware, and 2) computing pairwise distances and KDE within each training batch, with a time complexity of $\mathcal{O}(B^2d)$, where $B$ is the batch size (usually 256) and $d$ is the projection dimension (128 in our case). With DSA and IPC 50, it takes 55 seconds for 10 iterations without KDE and 65 seconds with KDE, resulting in an 18.2\% overhead. Note that, unlike SelecMix which updates the SupCon model during training and repeatedly recomputes pairwise distances, our pretrained SupCon model allows us to cache and reuse pairwise distances throughout the entire distillation process to mitigate this overhead. More details can be found in~\cref{appendix.time}.

 \section{Ablation Study}
\label{sec.ablation}
\textbf{1. Apply de-biasing methods to synthetic dataset.} 
Do we really need to mitigate bias in the DD process? Even though vanilla dataset distillation methods result in the biased synthetic set, can we still obtain an unbiased model from such a biased set using existing de-biasing training algorithms?  
\begin{table}
    \caption{Ablation study test accuracy (\%) on applying de-biasing method to train with synthetic datasets from DM, assessed under 5\% bias-conflicting samples and IPC 10 and 50.}
    \label{tab:debias-synthetic}
    \vskip 0.1in
    \centering
    \resizebox{1.0\linewidth}{!}{
    \begin{threeparttable}
    \begin{tabular}{ccccccccc}
    \toprule
     & \multicolumn{2}{c}{CMNIST} & \multicolumn{2}{c}{BG FMNIST}\\
     IPC & 10 & 50 & 10 & 50\\
     \midrule
     DM & 19.6$\pm$0.9 & 23.8$\pm$1.3 & 42.0$\pm$1.2 & 44.6$\pm$0.9\\
     DFA & 25.8$\pm$1.0 & 31.2$\pm$1.3 & 11.0$\pm$2.1 & 17.6$\pm$1.9 \\
     SelecMix\tnote{*} & \underline{43.3$\pm$1.3} & \underline{53.7$\pm$1.5} & \underline{57.2$\pm$1.1} & \underline{58.7$\pm$0.9}\\
     DM+Ours & \textbf{86.5$\pm$1.2} & \textbf{91.5$\pm$0.9} & \textbf{61.8$\pm$0.6} & \textbf{65.0$\pm$0.8}\\
    \bottomrule
    \end{tabular}
    \begin{tablenotes}
    \footnotesize
    \item [*] SelecMix has two versions; we choose the LfF-based~\cite{nam2020learning} version for its superior performance
    \end{tablenotes}
    \end{threeparttable}}
    \vskip -0.1in
\end{table}
In order to answer this question, we apply two SOTA de-biasing methods, SelectMix~\cite{hwangselecmix} and DFA~\cite{lee2021learning} on the synthetic dataset generated by DM and present the results in~\cref{tab:debias-synthetic}. The experiments are conducted on CMNIST and BG FMNIST with 5\% bias-conflicting samples and IPC 10 and 50. We have the following observations from the experiments: 1) DFA can slightly improve the performance on CMNIST but suffers from severe performance degradation on BG FMNIST. We think the reason is that DFA relies on a bias model to split embeddings, which suffers from performance degradation when the dataset becomes more complex, thus causing the overal performance drop. This aligns with the observations in~\cite{hwangselecmix}. 2)
While SelectMix consistently mitigates bias to some extent, it cannot fully rectify biases in synthetic datasets. For instance, SelectMix improves performance from 23.8\% to 53.7\% on CMNIST IPC 50. However, under the same bias conditions, standard training without dataset distillation achieves approximately $>$85\% accuracy. This indicates that even state-of-the-art methods struggle to recover from biases amplified synthetic sets. This observation aligns with \cref{fig:dm_vs_ours_ipc10}, which illustrates the severe bias amplification to the point where there isn't a single unbiased sample (in this case, a digit with a different color) in the synthetic set. Consequently, de-biasing becomes impossible in such a scenario. In contrast, our method achieves 91.5\% accuracy in this case, demonstrating the critical importance of debiasing during the dataset distillation procedure.\\
\textit{\textbf{Remark 1} While applying a state-of-the-art debiasing method typically enhances performance with a biased synthetic dataset, it still doesn't match up to the effectiveness of our KDE-based approach.}

\textbf{2. Applying de-biasing methods to surrogate models.} Many dataset distillation methods use surrogate models to distill information into synthetic sets. To mitigate bias, applying existing de-biasing methods to these surrogate models is another straightforward idea. Here, we explore the impact of applying de-biasing methods to surrogate models in DM and MTT. The experiments are conducted on CMNIST with 5\% bias conflicting sample. Since DM tries to match the distribution in the embedding space, we first apply DFA~\cite{lee2021learning} to separate the embedding into intrinsic (shape of the digits) and bias (color of the digits) parts.
\begin{table}
  \caption{Ablation study test accuracy (\%) on applying de-biasing methods to surrogate models on CMNIST with 5\% bias-conflict samples and IPC 1, 10 and 50.}
  \label{tab:debias-surrogate}
  \vskip 0.1in
  \centering
    \resizebox{0.9\linewidth}{!}{
  \begin{tabular}{cccc}
    \toprule
    & \multicolumn{3}{c}{IPC}                   \\
    Method     & 1     & 10 & 50 \\
    \midrule
    DM & 25.3$\pm$0.3 & 19.6$\pm$0.9 & 23.8$\pm$1.3\\
    DM+DFA & 26.1$\pm$0.3 &20.5$\pm$0.5&25.4$\pm$0.4 \\
    MTT & \underline{29.2$\pm$0.9} & \underline{47.7$\pm$0.8} & 33.9$\pm$1.2 \\
    MTT+SelecMix     & 18.1$\pm$0.5 & 29.9$\pm$0.8 & \underline{52.1$\pm$0.3}\\
    DM+Ours     & \textbf{32.2$\pm$1.0}&\textbf{86.5$\pm$1.2}&\textbf{91.5$\pm$0.9}  \\
    \bottomrule
  \end{tabular}}
  \vspace{-0.1in}
\end{table}
 Then we have DM match only the intrinsic part. For MTT, we first generate the de-biased expert training trajectories using SelecMix~\cite{hwangselecmix}. Then we have MTT match the de-biased training trajectories. The results are shown in Table \cref{tab:debias-surrogate}. We observe that there is a slight performance increase for DM+DFA compared to the vanilla DM which validates that the embeddings matched includes less biases. However, as SelecMix points out, fully separating intrinsic and bias parts is challenging. Thus biases will still be distilled into the synthetic dataset even if we only perform DM on the intrinsic embeddings.
For MTT+SelecMix, we see mixed results such as a 17.8\% drop on IPC 10 and an 18.2\% increase on IPC 50. We think the reason is that the de-biased expert training trajectories are not stable due to the use of auxiliary models. Although the final de-biased teacher model performs well, the intermediate training trajectories are hard for MTT to match.\\
\textit{\textbf{Remark 2} Our experiments reveal that biased features will still dominate even when using de-biased surrogate models during dataset distillation. 
This may be due to sub-optimal bias feature disentanglement and less stable matching trajectories.}

\begin{figure}
    \centering
\includegraphics[width=0.8\linewidth]{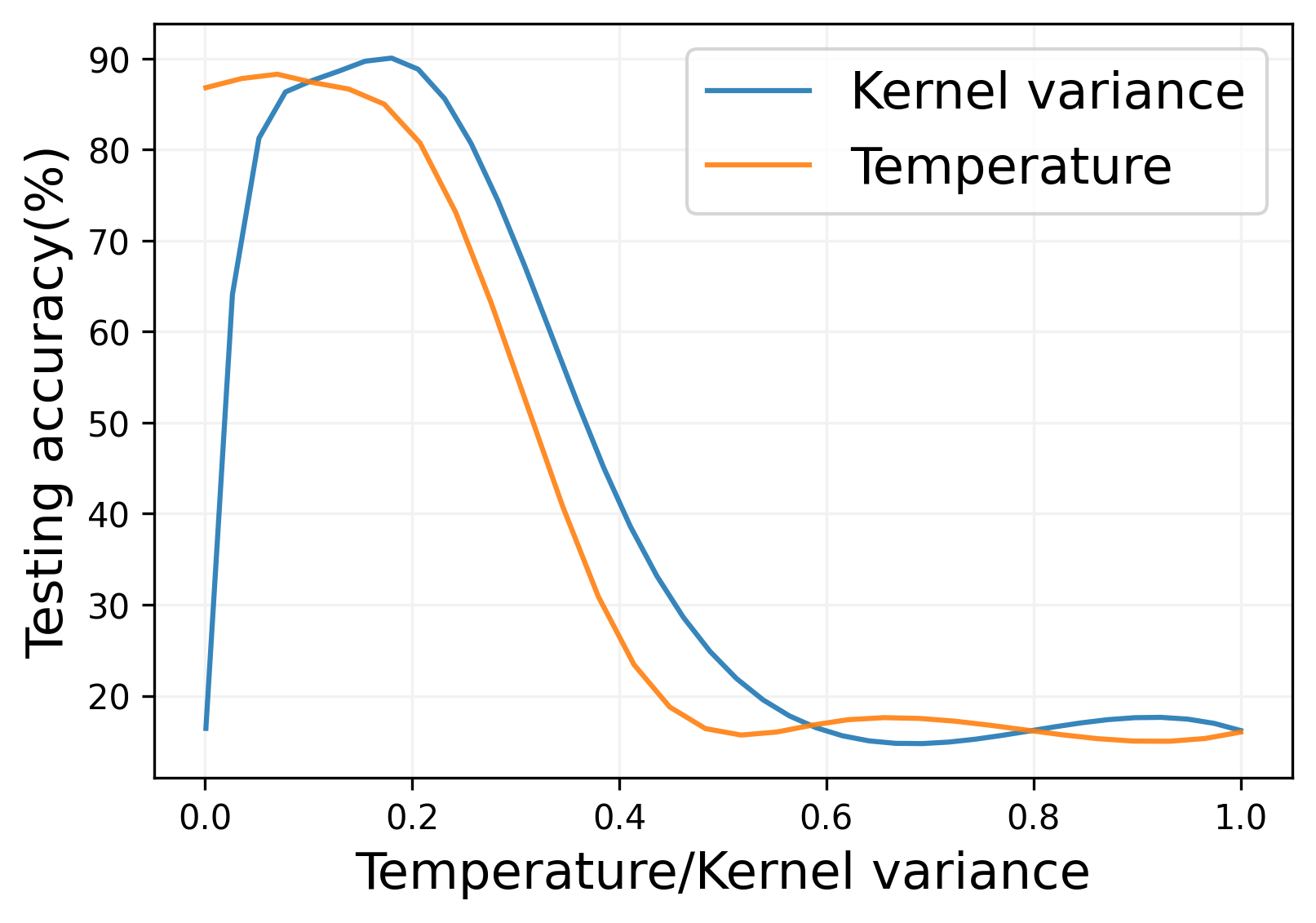}
    \caption{Ablation study on Kernel variance and temperature on CMNIST with 5\% bias-conflicting samples and IPC 10.}
    \label{fig:ablation-h-temp}
    \vspace{-0.15in}
\end{figure}

\textbf{3. Ablation study on hyper-parameters} We assess two hyper-parameters in our method, the kernel variance and temperature of $\mathbb{N}$, on CMNIST with 5\% bias-conflicting samples and IPC 10 and present the results in~\cref{fig:ablation-h-temp}. We use DM as the base method which achieves similar results but runs much faster than DSA.
In our observation, a very small variance introduces noise to the estimation, while a large value prevents the algorithm from assigning more weight to bias samples, leading to degraded performances. In general, choosing $\sigma^2$ to be $0.1$ works well, so we fix it in all of our experiments.

We also study the impact of softmax temperature when normalizing scores in~\cref{eq:kde}. When the temperature goes up, the weights are more evenly distributed among all samples. When the temperature goes down, more weights are given to samples that are further away from the rest of the samples. As shown in the figure, best performances are achieved around 0.1 which is the default setting.\\
\textit{\textbf{Remark 3} It can be seen that our method is robust to different hyper-parameter settings.}

\textbf{4. Ablation study on learning from de-biased dataset} Recent state-of-the-art de-biasing methods such as DFA~\cite{lee2021learning} and SelecMix~\cite{hwangselecmix} try to train the de-biased model by creating a less biased data representation. This leads us to wonder: Can we follow the same procedure and have DD methods train on de-biased representations generated by these methods? In this ablation study, we run DM on de-biased dataset generated by one of the recent SOTA methods SelecMix and show the results in \cref{table:ablation_unbiased_dataset}. The experiment is conducted on CMNIST with IPC 10 and varying bias-conflict ratios.
\begin{table}
    \caption{Ablation study test accuracy (\%) on creating de-biased dataset first using SelecMix and then applying dataset distillation method on it. The experiment is run on CMNIST with IPC 10.}
    \label{table:ablation_unbiased_dataset}
    \vskip 0.1in
    \centering
    \resizebox{0.95\linewidth}{!}{
    \begin{tabular}{ccccccccc}
    \toprule
     & \multicolumn{3}{c}{Bias-conflict Ratio} \\
     Method & 1\% & 2\% & 5\%\\
     \midrule
     DM& 18.6$\pm$0.2& 18.5$\pm$0.6& 19.6$\pm$0.9\\
     SelecMix+DM&\underline{44.6$\pm$1.0}&\underline{25.6$\pm$0.9}&\underline{65.6$\pm$0.8}\\
     DM+Ours&\textbf{64.9$\pm$0.3}&\textbf{50.6$\pm$1.2}&\textbf{86.5$\pm$1.2}\\
    \bottomrule
    \end{tabular}
    }
\end{table}
As we can see from the table that the performances are greatly boosted after creating a less biased representation and have DM learn from it. This aligns with our earlier observation that dataset distillation methods can effectively learn from unbiased datasets. However, its performance still falls short compared to our KDE based method. We think the reason is that KDE estimates a smoothed distribution of the biased and unbiased samples while SelecMix combines two data points with lowest similarity (e.g. intra-class) which depends heavily on the auxiliary model quality.\\
\textit{\textbf{Remark 4} Creating an unbiased dataset first and then applying dataset distillation methods on it also greatly boost the model performance. However it still falls short by a large margin compared to our method.}

\section{Conclusion}
This paper conducts the first analysis of dataset bias in dataset distillation. Our findings show that bias type greatly influences distillation behavior (amplification vs. suppression). Then we introduce a debiasing method using re-weight and kernel density estimation which substantially reduces retained bias in distilled datasets. We assess our debiasing method on various benchmark datasets with different bias ratios and IPC values and empirically verify the effectiveness of our method. 
In summary, our study offers insights into bias in dataset distillation, presents a practical algorithm for better performance and paves the way for future research on bias mitigation.

\textbf{Limitations.} Although many dataset distillation methods rely on the matching of real and synthetic dataset through carefully designed objective functions, there are methods such as MTT where our method cannot be easily applied due to the use of expert training trajectories instead of real images directly. This will be the focus of our future research.

\section*{Impact Statement} 
Our paper will have the following positive social impacts: 1) We are the first revealing bias amplification effect in data distillation processes, which can prevent the misuse of such technique. 2) We propose a simple yet effective way to mitigate bias when data distillation is used, which can lead to unbiased decisions when employing those models. 

\section*{Acknowledgements}
The work is partially supported by NSF 2048280, 2331966, 2325121, 2244760 and ONR N00014-23-1-2300.

\nocite{langley00}

\bibliography{example_paper}
\bibliographystyle{icml2024}

\newpage
\appendix
\onecolumn
\section{Appendix}

\subsection{Extension to Other SOTA methods}
\label{appendix.extension}
Here we show that our method can be extended to other SOTA methods such as DREAM~\cite{liu2023dream} and IDM~\cite{zhao2023improved} that's based on either DM or DSA. In DREAM, instead of sampling and matching random batches of real images, it applies Kmeans~\cite{lloyd1982least} algorithm to select representative samples. However, if a dataset is dominated by biased samples, this will cause the bias to be amplified, thus making de-biasing harder. Same for IDM which applies extra operations on top of DM. Both DREAM and IDM shares similar loss with DSA and DM with extra loss to capture the information shared between the condensed dataset and the real dataset. Therefore, our method can naturally extend to these methods as well. See~\cref{table:extension_to_dream} for the detailed performance boost.
\begin{table*}[h]
  \centering
  \caption{Extend our method to recent SOTA methods DREAM and IDM. Results are run on CMNIST dataset with various bias-conflict ratios.}
  \label{table:extension_to_dream}
  \vskip 0.15in
  \resizebox{1.0\textwidth}{!}{
  \begin{threeparttable}
  \begin{tabular}{ccccccccccc}
    \toprule
    Method & \multicolumn{3}{c}{Bias-conflict Ratio (1.0\%)} &\multicolumn{3}{c}{Bias-conflict Ratio (2.0\%)} & \multicolumn{3}{c}{Bias-conflict Ratio (5.0\%)} \\
    &  1 & 10 & 50 & 1 & 10 & 50 & 1 & 10 & 50\\
    \midrule
    
    IDM & 20.1$\pm$0.9&18.8$\pm$1.2&17.6$\pm$1.6 &
    20.2$\pm$1.0&20.1$\pm$0.5&19.1$\pm$1.3 & 
    20.8$\pm$0.9&22.3$\pm$1.1&26.5$\pm$1.5\\
    IDM+Ours & 26.1$\pm$1.0&54.1$\pm$0.3&72.2$\pm$0.1&27.8$\pm$1.0&35.2$\pm$0.1&55.7$\pm$0.1&30.9$\pm$0.5&40.4$\pm$0.4&77.2$\pm$0.2\\
    DREAM & 21.3$\pm$0.4& 15.9$\pm$1.2 & 17.9$\pm$0.8 & 26.2$\pm$0.5 & 17.1$\pm$1.3 & 17.5$\pm$1.5 & 23.7$\pm$1.2 & 30.8$\pm$1.5 & 50.4$\pm$1.0\\
    DREAM+Ours & 27.0$\pm$1.2&56.6$\pm$0.8&73.8$\pm$1.1&27.4$\pm$1.1&38.3$\pm$1.1&58.5$\pm$0.1&40.1$\pm$0.5&50.0$\pm$1.2&80.1$\pm$0.6\\
    \bottomrule
  \end{tabular}
  \end{threeparttable}
  }
\end{table*}

\subsection{Experimental Setup}
\label{appendix.exp_setup}
Following previous dataset distillation methods~\cite{zhaodm,cazenavette2022dataset}, we use ConvNet as the model architecture. It has 128 filters with kernel size of 3$\times$3. Then it's followed by instance normalization, RELU activation, and an average pooling layer. We use SGD as the optimizer with 0.01 learning rate. For the supervised contrastive model, we use ResNet18~\cite{he2016deep} following~\cite{hwangselecmix} with a projection head of 128 dimensions. Same as previous de-biasing works~\cite{nam2020learning,lee2021learning,hwangselecmix}, we evaluate our results by training a DNN model (same as distillation) on the synthetic dataset and measure its accuracy using the unbiased test set. The results are evaluated with IPC 1, 10 and 50. See \cref{appendix.exp_setup} in Appendix for more experiment details.

To avoid introducing biases during initialization, all synthetic images are initialized with random noise instead of randomly selected real images. For KDE, we fix kernel variance and temperature to be 0.1 across all datasets. All experiments are run on a single 48GB NVIDIA RTX A6000 GPU.

\subsection{Extremely low bias-conflict setting}
\label{appendix.extreme}
It's possible that in some datasets that bias-conflict samples are extremely rare such as 0.5\% used in SelecMix~\cite{hwangselecmix}. Here we show that our method performs well under this setting as well. See~\cref{table:appendix.extreme}.
\begin{table}[h]
  \caption{Test accuracy under extremely low bias-conflict ratios (0.5\%).}
  \label{table:appendix.extreme}
  \vskip 0.15in
  \centering
  \resizebox{0.8\linewidth}{!}{
  \begin{threeparttable}
  \begin{tabular}{ccccccc}
    \toprule
    Dataset & IPC &  MTT &  DM & DM+Ours & DSA & DSA+Ours \\
    \midrule
    \multirow{3}{*}{CMNIST} &1 & 22.9$\pm$1.2 & 25.1$\pm$0.9&27.4$\pm$0.7&25.9$\pm$0.7&26.0$\pm$1.0\\
    &10 & 26.8$\pm$1.8&18.8$\pm$0.9&40.0$\pm$1.2&15.7$\pm$0.8&42.0$\pm$1.2\\
    &50& 20.7$\pm$1.0&23.8$\pm$0.9&49.0$\pm$1.0&11.6$\pm$2.0&50.0$\pm$1.5\\
    \bottomrule
  \end{tabular}
    \end{threeparttable}
  }
\end{table}

\subsection{Experiment results on the original unbiased dataset}
Here we show the experiment results of our method on the original unbiased dataset in~\cref{appendix:tab:original_dataset}.
\begin{table}[ht]
\centering
\caption{Performance Comparison on original Datasets}
\label{appendix:tab:original_dataset}
\begin{tabular}{llccc}
\toprule
Dataset       & Method   & IPC 1           & IPC 10          & IPC 50          \\ \midrule
\multirow{2}{*}{CMNIST}        & DM       & $87.3\pm0.4$    & $95.6\pm0.5$    & $97.2\pm0.7$    \\
              & DM+Ours  & $86.0\pm0.5$    & $95.8\pm0.8$    & $97.2\pm1.0$    \\
              \midrule
\multirow{2}{*}{FashionMNIST} & DM       & $72.5\pm0.5$    & $83.1\pm0.4$    & $86.2\pm1.0$    \\
              & DM+Ours  & $70.4\pm0.9$    & $82.4\pm0.7$    & $84.5\pm1.2$    \\
              \midrule
\multirow{2}{*}{CIFAR-10}     & DM       & $27.2\pm0.3$    & $49.5\pm0.6$    & $60.2\pm0.8$    \\
              & DM+Ours  & $27.1\pm0.2$    & $49.0\pm0.8$    & $60.8\pm0.9$    \\ \bottomrule
\end{tabular}
\end{table}

\subsection{Datasets}
\label{appendix.sec.datasets}
We conduct experiments on 3 datasets,  two of which are widely used in SOTA de-biasing methods to assess color and noise biases, while also introducing a novel dataset to evaluate background biases.

\textbf{Colored MNIST (CMNIST):} \cite{nam2020learning} introduces the Colored MNIST dataset by injecting color with random perturbation into the MNIST dataset~\cite{deng2012mnist}. Each digit will be associated with a specific color as its bias such as digit 0 being red and digit 4 being green. We evaluate our method under 3 bias conflicting ratios with 1\%~(54,509 bias aligned, 491 bias conflicting), 2\%~(54,014 bias aligned, 986 bias conflicting) and 5\%~(52,551 bias aligned, 2,449 bias conflicting).

\textbf{Corrupted CIFAR-10:} Generated from the regular CIFAR-10 dataset~\cite{krizhevsky2009learning}, Corrupted CIFAR-10 applies different corruptions~\cite{Hendrycks_Dietterich_2019} to the images in CIFAR-10 so that images from one class are associated with one type of corruption such as GaussianNoise or MotionBlur. We also evaluate our method under 3 bias conflicting ratios with 1\%~(44,527 bias aligned, 442 bias conflicting), 2\%~(44,145 bias aligned, 887 bias conflicting) and 5\%~(42,820 bias aligned, 2,242 bias conflicting).

\textbf{Background Fashion-MNIST (BG FMNIST):} 
Background bias, which results in an over-reliance on the background for predicting foreground objects, has been employed to assess interpretability methods in various prior studies~\cite{BAM2019,zhou2017places}.
Following this idea, we construct a new dataset biased in backgrounds by using Fashion-MNIST~\cite{xiao2017fashionmnist} as foregrounds which include a training set of 60,000 examples and a test set of 10,000 examples. And MiniPlaces~\cite{zhou2017places} is used as backgrounds to introduce background biases such as T-shirt is associated with bamboo forest background, trouser is associated with livingroom background, etc. Similar to other biased datasets, we also conduct experiments in 3 settings including 1\%, 2\% and 5\% bias conflicting samples.

\textbf{BFFHQ:} It is introduced in~\cite{lee2021learning} by curating a biased dataset from the FFHQ dataset~\cite{karras2019style}. In this dataset, they focus on age and gender as key facial attributes, with age being the intrinsic attribute and gender as the bias attribute. The dataset is curated to exhibit a strong correlation between these two attributes. The dataset predominantly features females categorized as 'young' (aged between 10 to 29 years) and males as 'old' (aged between 40 to 59 years). Consequently, the majority of the samples in the dataset are composed of young women and old men, aligning with the intended bias. The dataset only comes with 0.5\% bias-conflict samples. For more detailed info regarding the dataset, please refer to~\cite{lee2021learning}.

\subsection{Matching Training Trajectories}
The objective function of MTT can be described as below
\begin{equation}
    \label{eq:tm_loss}
    \mathcal{L} = \| \hat{\theta}_{t+T} - \theta_{t+M}^* \|^2_2 / \| \theta_t^{*} - \theta_{t+M}^* \|_2^2.
\end{equation}
Where T is the number of synthetic data training steps and M is the expert model training steps using the whole dataset. As MTT tries to match the training trajectory of expert models and models trained using synthetic data, the real dataset is not needed during matching phase. Therefore, our reweighting scheme doesn't work directly in MTT. However, as shown in~\cref{tab:debias-surrogate}, we should still be able to mitigate bias for MTT through fixing the expert models. We leave this to future work.

\subsection{Qualitative Analysis}
\label{appendix.qa}
\begin{figure}
    \centering
    \includegraphics[width=0.48\linewidth]{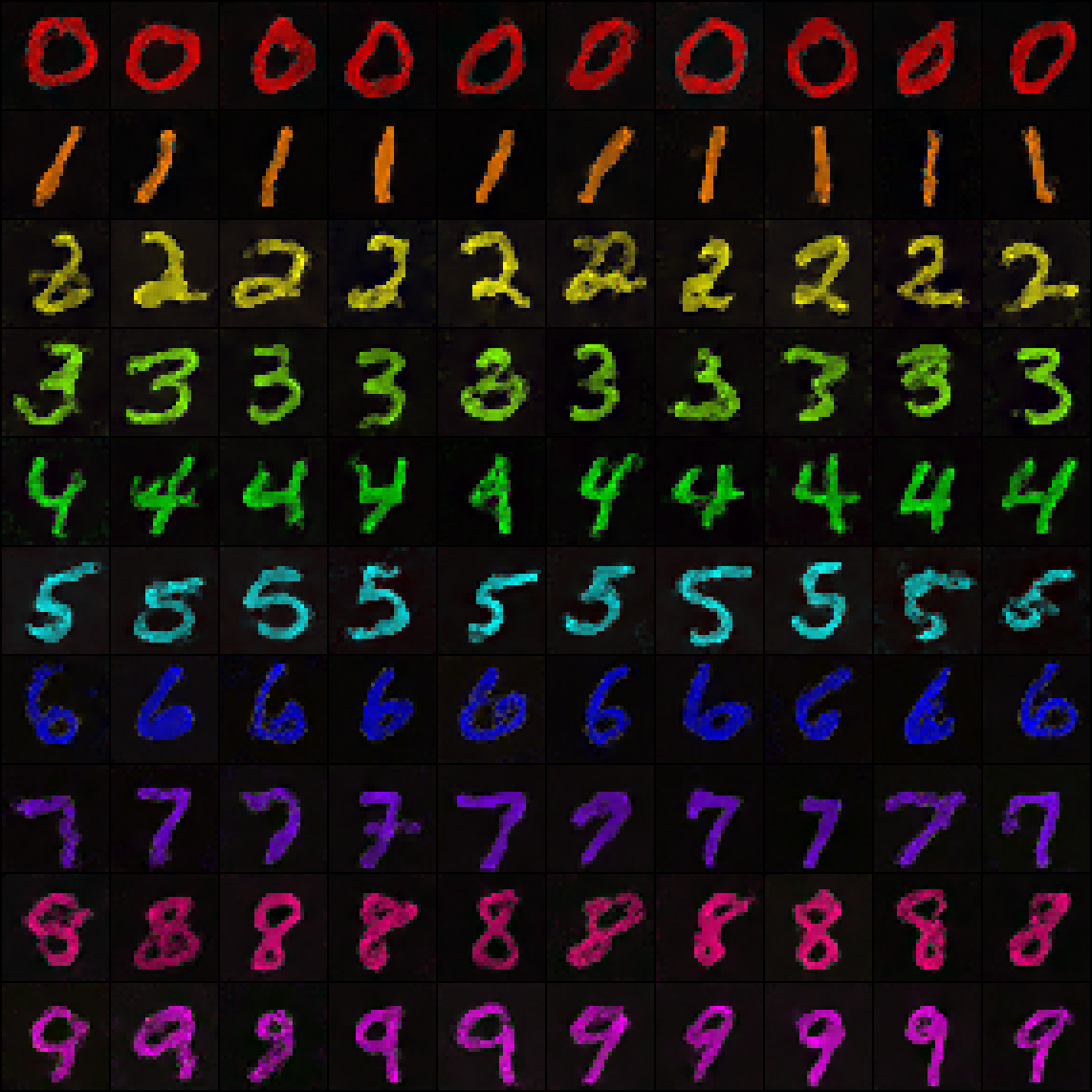}
    \includegraphics[width=0.48\linewidth]{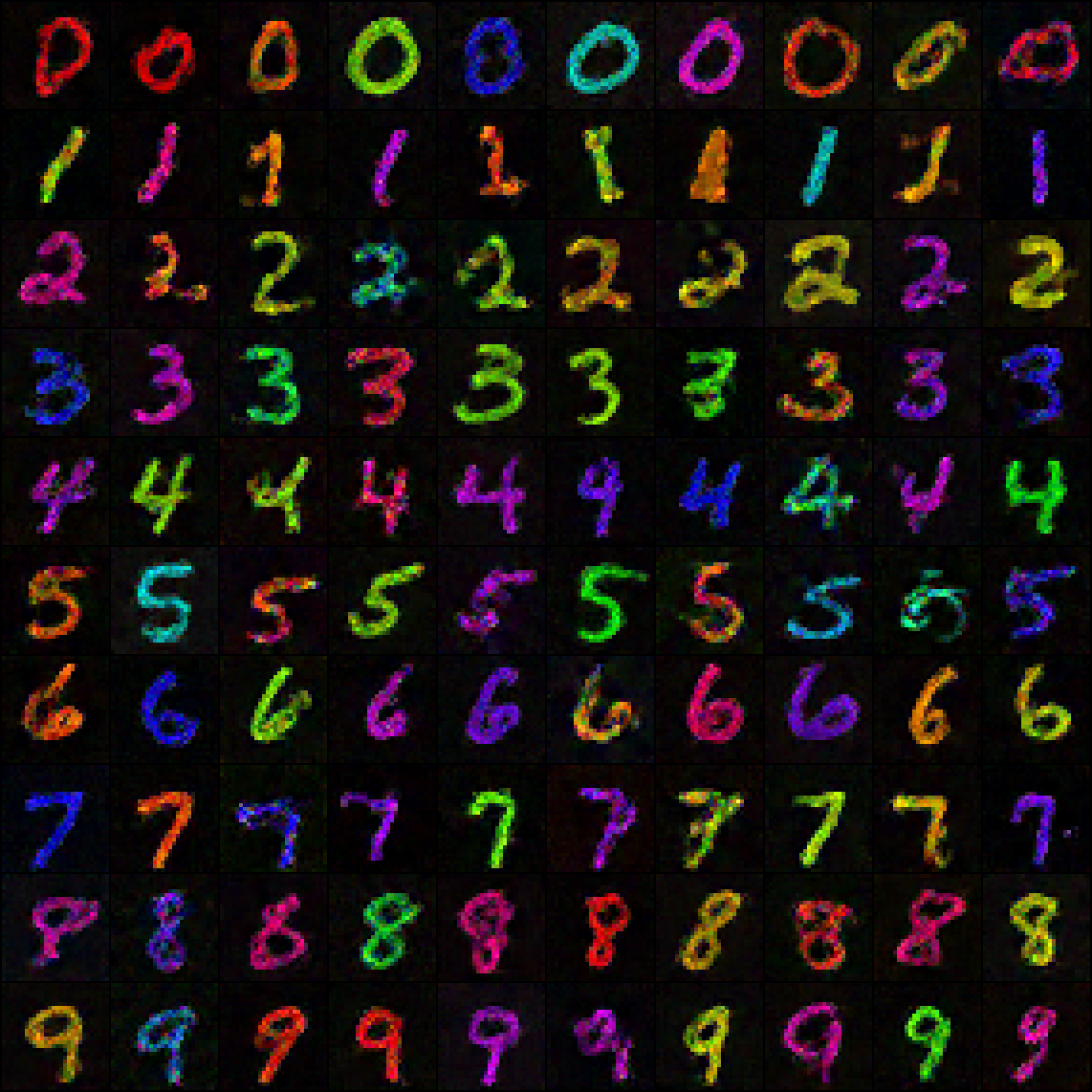}
    \caption{Synthetic images from vanilla DM (left) vs Ours (right) distilled from CMNIST with 5\% bias-conflict samples. The one synthesized by vanilla DM is dominated by the bias feature while ours includes a rich set of features from both biased and bias-conflict samples.}
    \label{fig:dm_vs_ours_ipc10}
\end{figure}
Here we visualize the synthetic datasets produced by vanilla DM and DM+Ours on CMNIST with 5\% bias-conflicting samples and IPC 10 in~\cref{fig:dm_vs_ours_ipc10}.  
 As shown, images synthesized by the vanilla DM completely ignores the unbiased samples due to the reasons explained in ~\cref{sec.dm_ours}, causing the bias to be even more amplified than the original dataset (the original dataset has 5\% unbiased samples such as green 0s or yellow 1s, the distilled dataset has 0\%). When combined with our method, DM is able to identify and synthesize unbiased samples into the final synthetic dataset. Similar results can also be seen with the vanilla DSA and DSA+Ours in Appendix~\cref{appendix.sample_images}. 

\subsection{Visualization of KDE}
\begin{figure}
    \centering
    \includegraphics[width=0.5\textwidth]{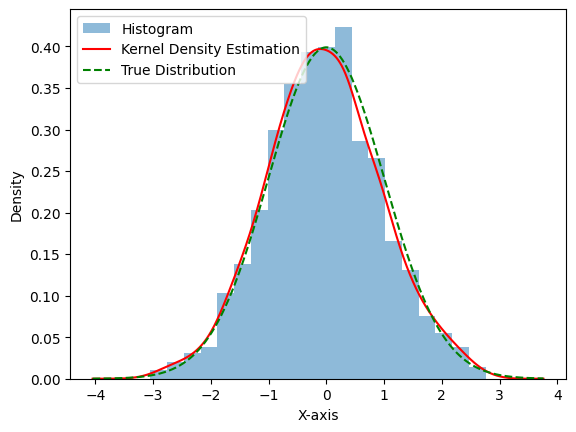}
    \caption{KDE applied on a normal distribution.}
    \label{fig:kde-normal}
\end{figure}

See Figure \cref{fig:kde-normal} for a visualization of KDE applied on a normal distribution. The dotted line is a true normal distribution. The histogram represents the observed data points and the red curve shows the density function estimated using KDE.

\subsection{Ablation study on cutting-off score}
\begin{figure}
    \centering
    \includegraphics[width=0.5\textwidth]{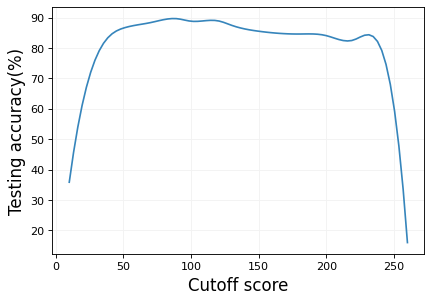}
    \caption{Ablation study on cutoff score. The experiment is conducted on CMNIST with 5\% bias-conflicting samples and IPC 10.}
    \label{fig:ablation-cutoff-score}
\end{figure}

Here we conduct an ablation study on the cutoff score used when computing the sample weights, the results are shown in~\cref{fig:ablation-cutoff-score}. It can be seen that choosing a cutoff score that's too large or too small will hinder the de-biasing performance. Also our algorithm is not very sensitive to cutoff scores and the optimal performances can be achived with a wide range of cutoff scores.

\subsection{Implementation details of applying de-biasing methods to synthetic dataset}
As mentioned before, one natural idea is to directly apply de-biasing methods on distilled synthetic dataset. Here we describe the implementation details. For the results of DM, it's acquired by directly running the vanilla DM on CMNIST dataset. For DM+DFA, we apply the DFA de-biasing methods on the distilled synthetic dataset from the original DM. For DM+SelecMix, we test the two methods proposed in~\cite{hwangselecmix} which corresponds to directly applying SelecMix methods and appling it on top of LfF~\cite{nam2020learning}.

\subsection{Implementation details of applying de-biasing method to surrogate model}
For applying de-biasing methods to surrogate models, we mainly test two combinations. The first one is DM plus DFA. The reason this can potentially work is that DM tries to match the distribution of real data and synthetic data in the embedding space. DFA is a perfect fit for DM because DFA tries to separate the embeddings into the intrinsic parts and the bias parts. DM can choose to match only the intrinsic parts, thus getting rid of biases. For MTT, since it doesn't rely on real data during matching phase but the expert training trajectories. Therefore, the best way to mitigate bias in MTT synthesized datasets is to debias the expert training trajectories. Thus we choose to apply the most recent SOTA model de-biasing method to acquire the expert trajectories first. Then we have MTT match these de-biased trajectories. 

\subsection{Computation time for KDE}
\label{appendix.time}
Similar to~\cite{hwangselecmix}, computing the embedding distance between samples adds extra computational cost. This applies to our method as well. For distribution based method such as DM, since it doesn't compute second order gradients, the whole distillation process is very fast. The time used to compute 10 matching iterations on CMNIST for the vanilla DM is 1.0 second while the time used to compute 10 matching iterations of DM+KDE is 1.9 seconds. However, note that DM usually takes less than 10,000 iterations to converge which is around 16.7 minutes. Compared to other dataset distillation based methods such as DSA that takes hours to finish on the same dataset, DM+KDE can converge quickly even with the added computation. For DSA, sicne it computes second order gradients, it takes around 55 seconds for each 10 iterations. With KDE, it takes 65 seconds. Therefore, computing KDE adds around 18.2\% computation time. We usually observe significant performance improvement at the early training epochs when combining our method with the vanilla method. Note that generating the Supervised Contrastive model is fast, e.g. it takes less than 30 minutes to train it on CMNIST. All the data are measured on the same hardware as our experiments.

The computation of KDE can be greatly reduced by caching or grouping the data points. Based on our findings mentioned in the main paper, we observe that most of the biased samples tend to be close to each other in the SupCon embedding space and most of the unbiased samples are far away from this majority group. Therefore, there is no need to compute their distance within these biased samples and their distance with unbiased samples repeatedly. Thus, by pre-computing an estimated grouping information and reuse the intra-group and inter-group distance, the whole KDE process can be greatly reduced. We leave this to our future work.

\begin{figure*}
    \centering
    \includegraphics[width=0.32\linewidth]{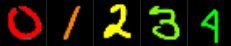}
    \includegraphics[width=0.32\linewidth]{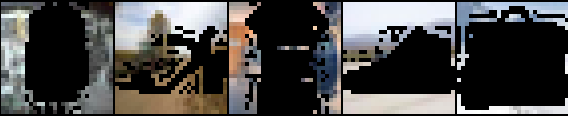}
    \includegraphics[width=0.32\linewidth]{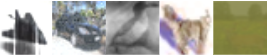}
    \includegraphics[width=0.32\linewidth]{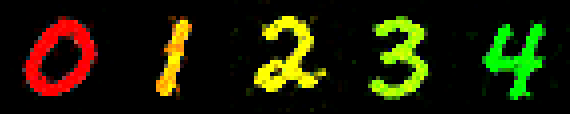}
    \includegraphics[width=0.32\linewidth]{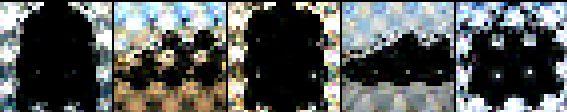}
    \includegraphics[width=0.32\linewidth]{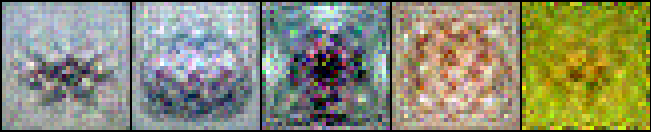}
    \caption{Examples of original images (top) and synthetic images (bottom) from biased dataset.}
    \label{fig:example_images}
\end{figure*}

\subsection{More qualitative results}
\label{appendix.qualitative}
\begin{figure}[h]
     \centering
     \begin{subfigure}[b]{0.48\textwidth}
        \centering
        \includegraphics[width=\linewidth]{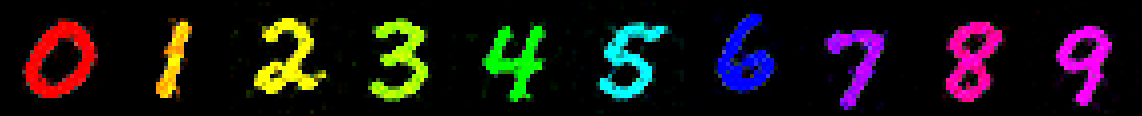}
        \includegraphics[width=\linewidth]{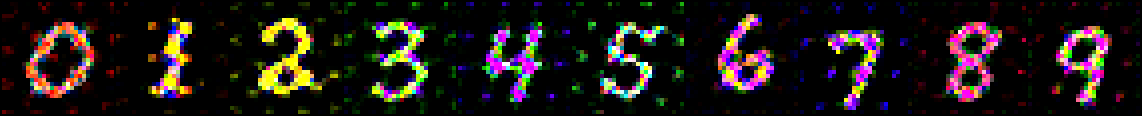}
        \label{fig:dm_vs_ours_ipc1}
     \end{subfigure}
     \hfill
     \begin{subfigure}[b]{0.48\textwidth}
         \centering
          \includegraphics[width=1.0\textwidth]{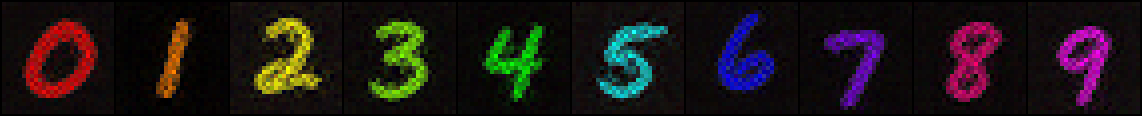}
        \includegraphics[width=1.0\textwidth]{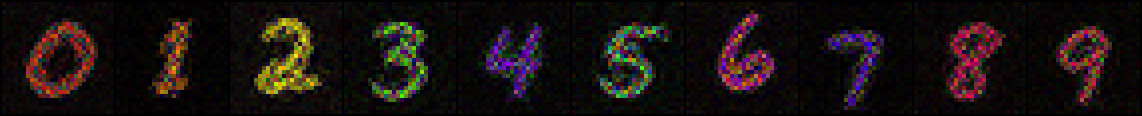}
    \label{fig:dsa_vs_ours_ipc1}
     \end{subfigure}
     \caption{Synthetic images from original DM (top left) vs DM+Ours (bottom left) and original DSA (top right) vs DSA+Ours (bottom right). Results are generated from CMNIST with 5\% bias-conflicting samples and IPC 1. }
\end{figure}

First of all, we show the results under IPC 1 in~\cref{fig:dm_vs_ours_ipc1}. It can be seen that images synthesized by the vanilla DM is dominated by the biased samples such as red 0s and green 4s. On the contrary, the ones produced by our method is a fusion of both biased samples and unbiased samples, thus mitigating the biases in synthetic datasets. Similar results can also be seen from gradient based method in~\cref{fig:dsa_vs_ours_ipc1}
where the single image synthesized by DSA is also a fusion of both biased and unbiased samples.

\subsection{Example Synthetic Images}
\label{appendix.sample_images}
\begin{figure*}
    \centering
    \includegraphics[width=\textwidth]{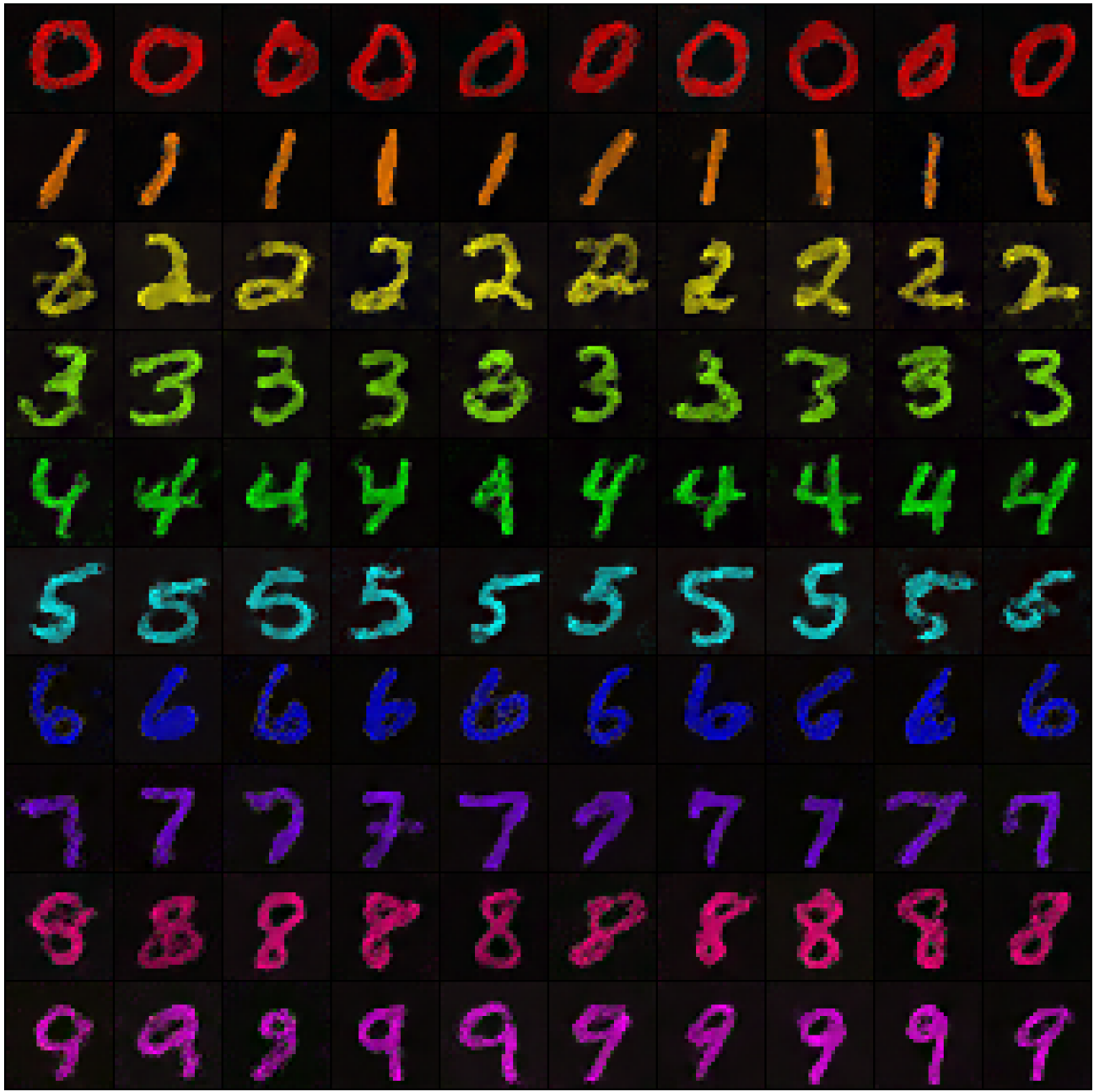}
    \caption{Synthesized dataset by DM on CMNIST with 5\% bias-conflicting samples.}
\end{figure*}

\begin{figure*}
    \centering
    \includegraphics[width=\textwidth]{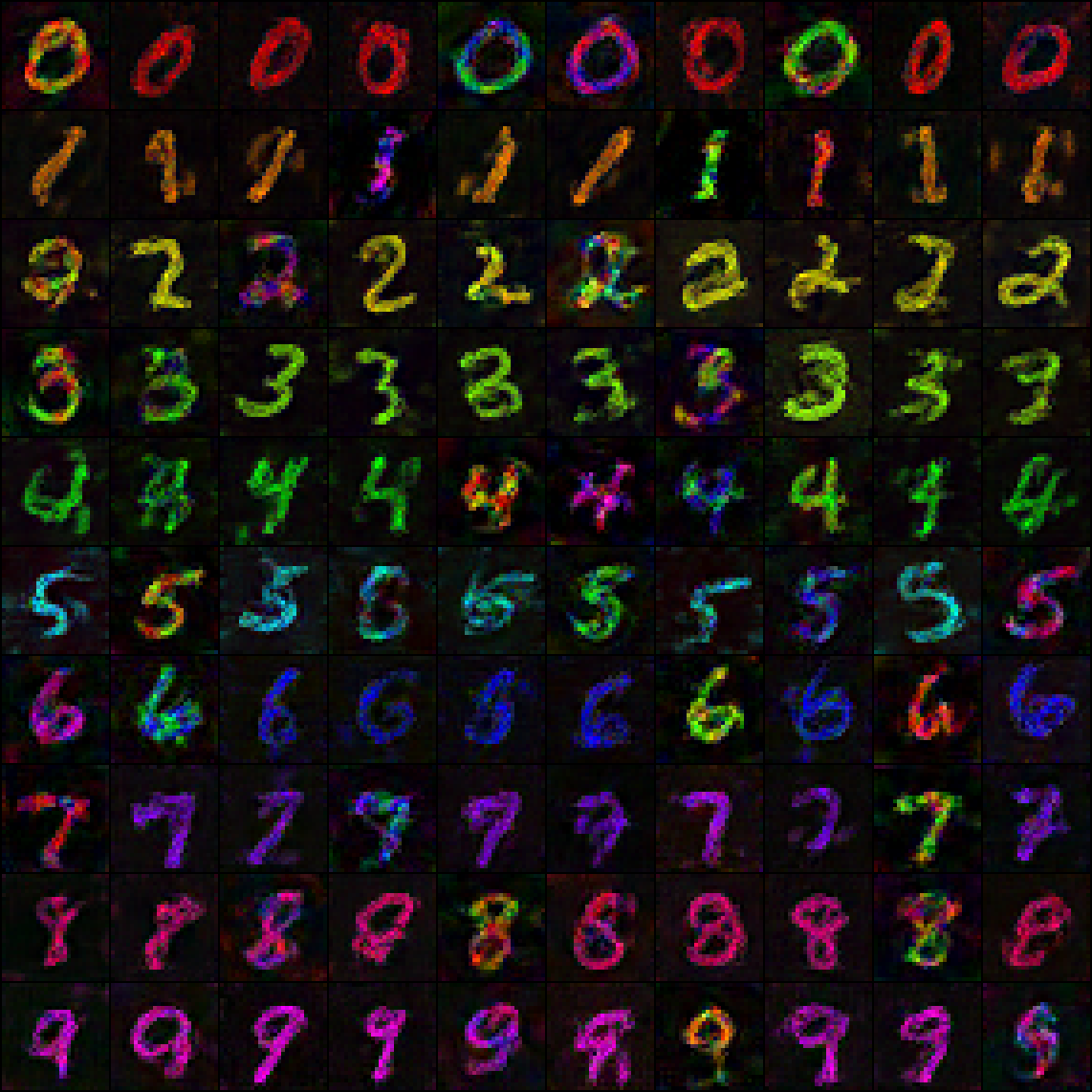}
    \caption{Synthesized dataset by DSA on CMNIST with 5\% bias-conflicting samples.}
\end{figure*}

\begin{figure*}
    \centering
    \includegraphics[width=\textwidth]{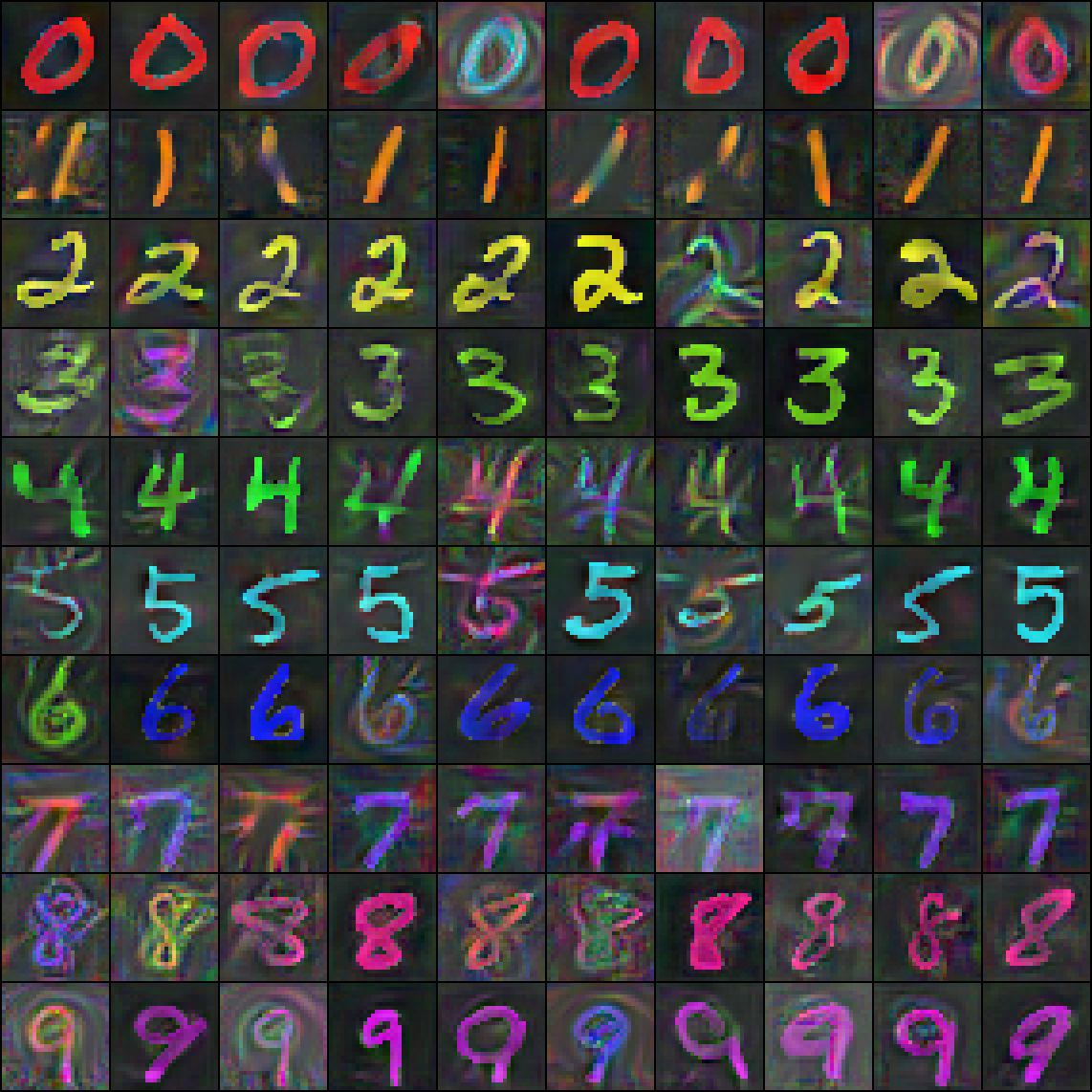}
    \caption{Synthesized dataset by MTT on CMNIST with 5\% bias-conflicting samples.}
\end{figure*}

\begin{figure*}
    \centering
    \includegraphics[width=\textwidth]{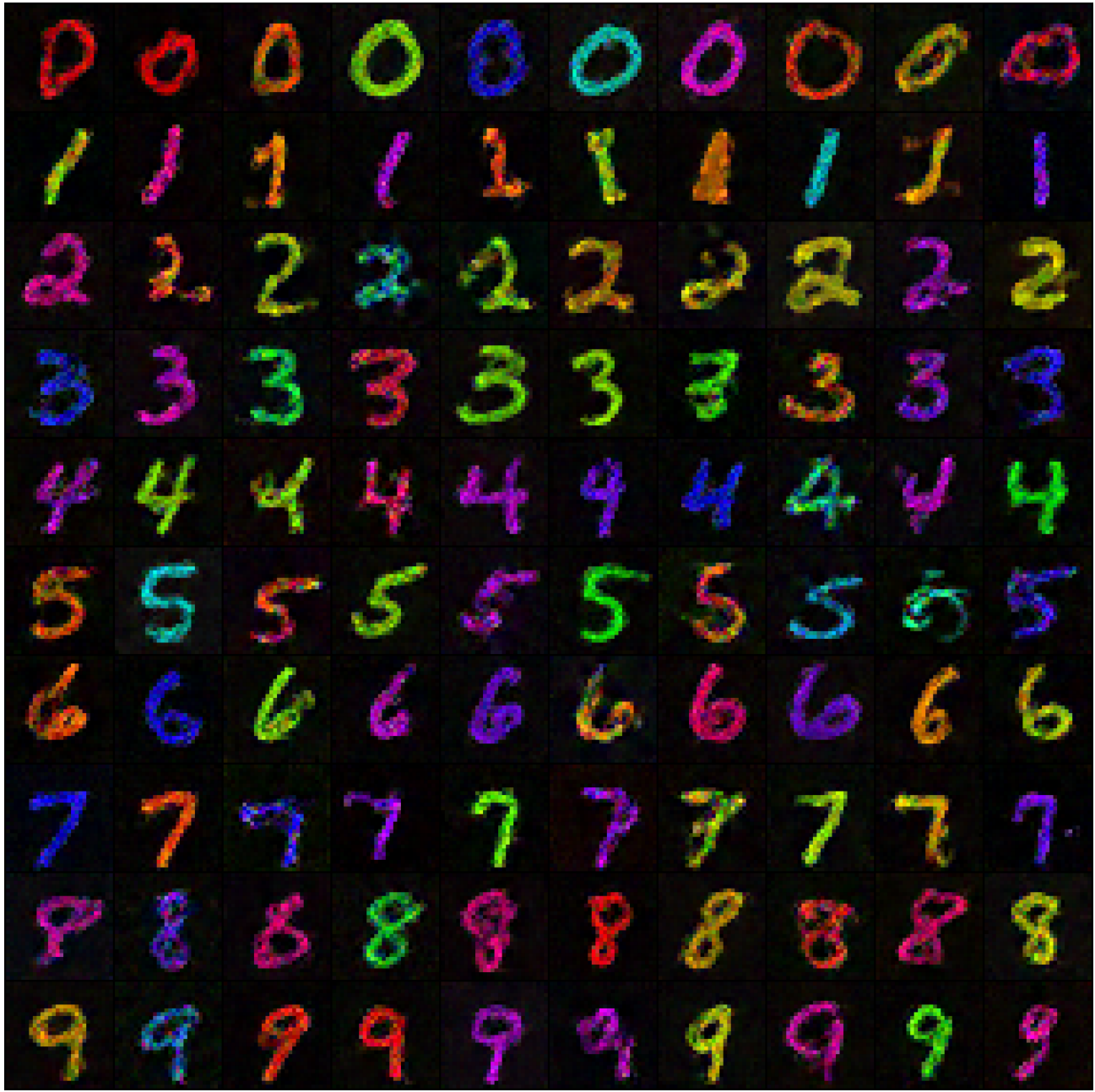}
    \caption{Synthesized dataset by DM+Ours on CMNIST with 5\% bias-conflicting samples.}
\end{figure*}

\begin{figure*}
    \centering
    \includegraphics[width=\textwidth]{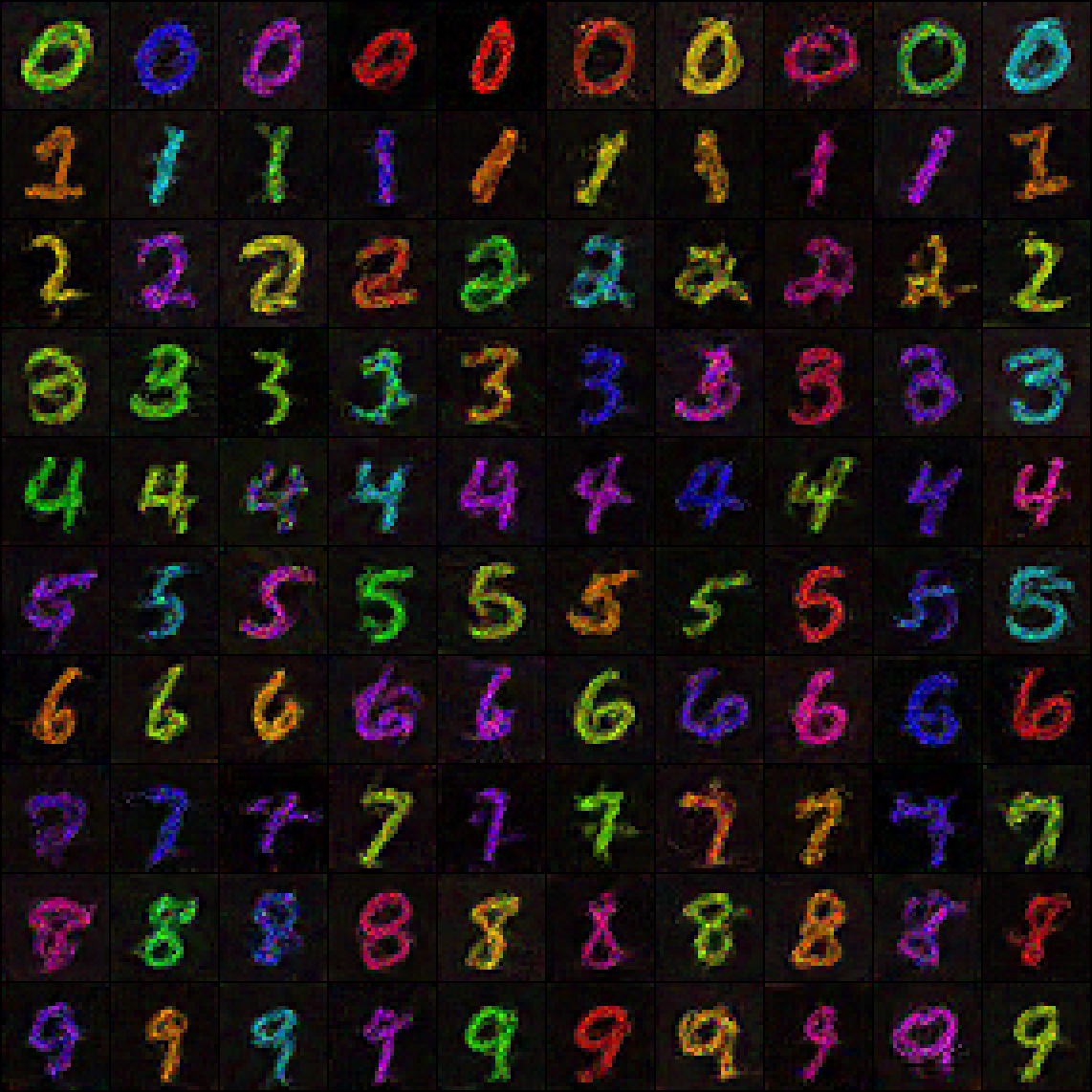}
    \caption{Synthesized dataset by DSA+Ours on CMNIST with 5\% bias-conflicting samples.}
\end{figure*}

\begin{figure*}
    \centering
    \includegraphics[width=\textwidth]{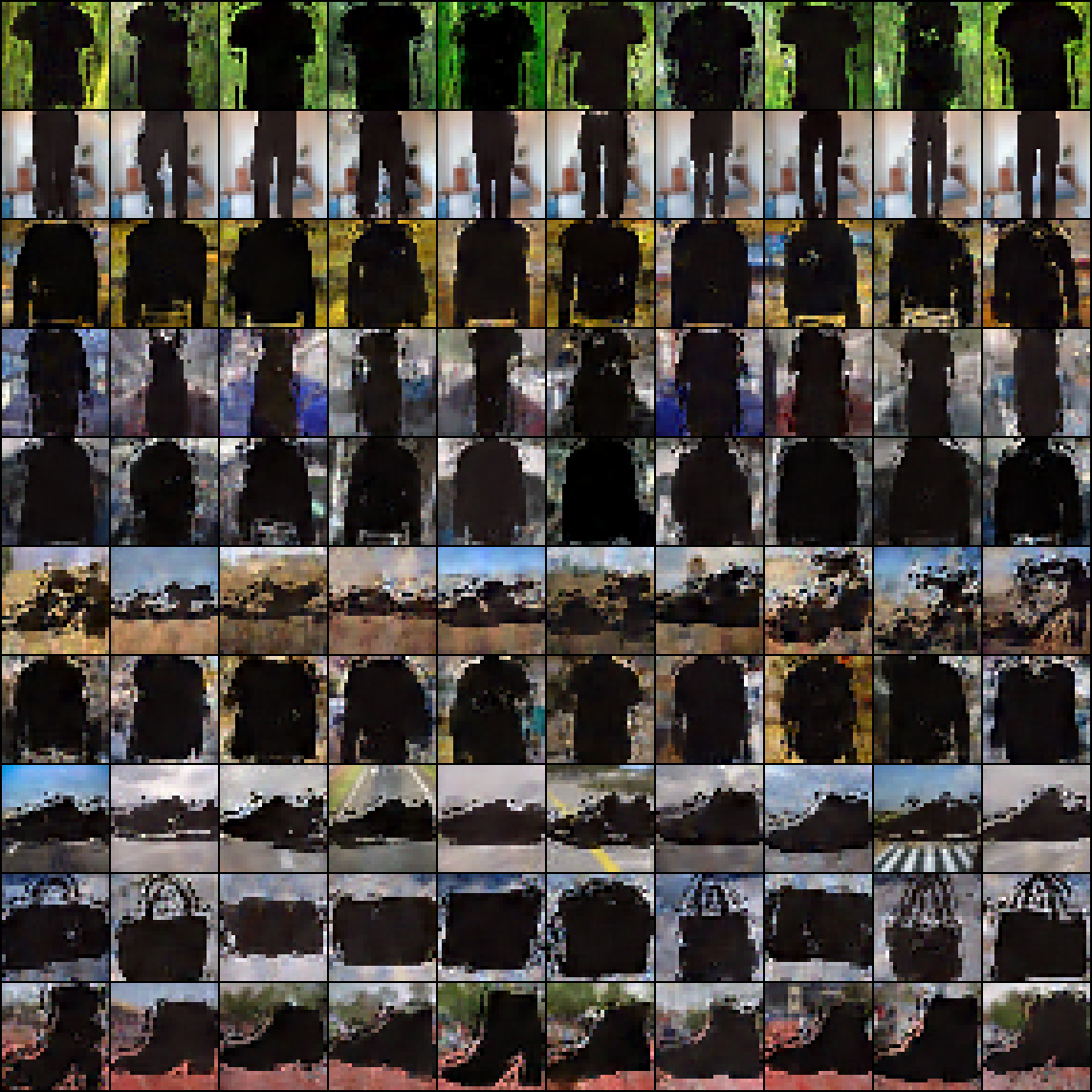}
    \caption{Synthesized dataset by DM on BG FMNIST with 5\% bias-conflicting samples.}
\end{figure*}

\begin{figure*}
    \centering
    \includegraphics[width=\textwidth]{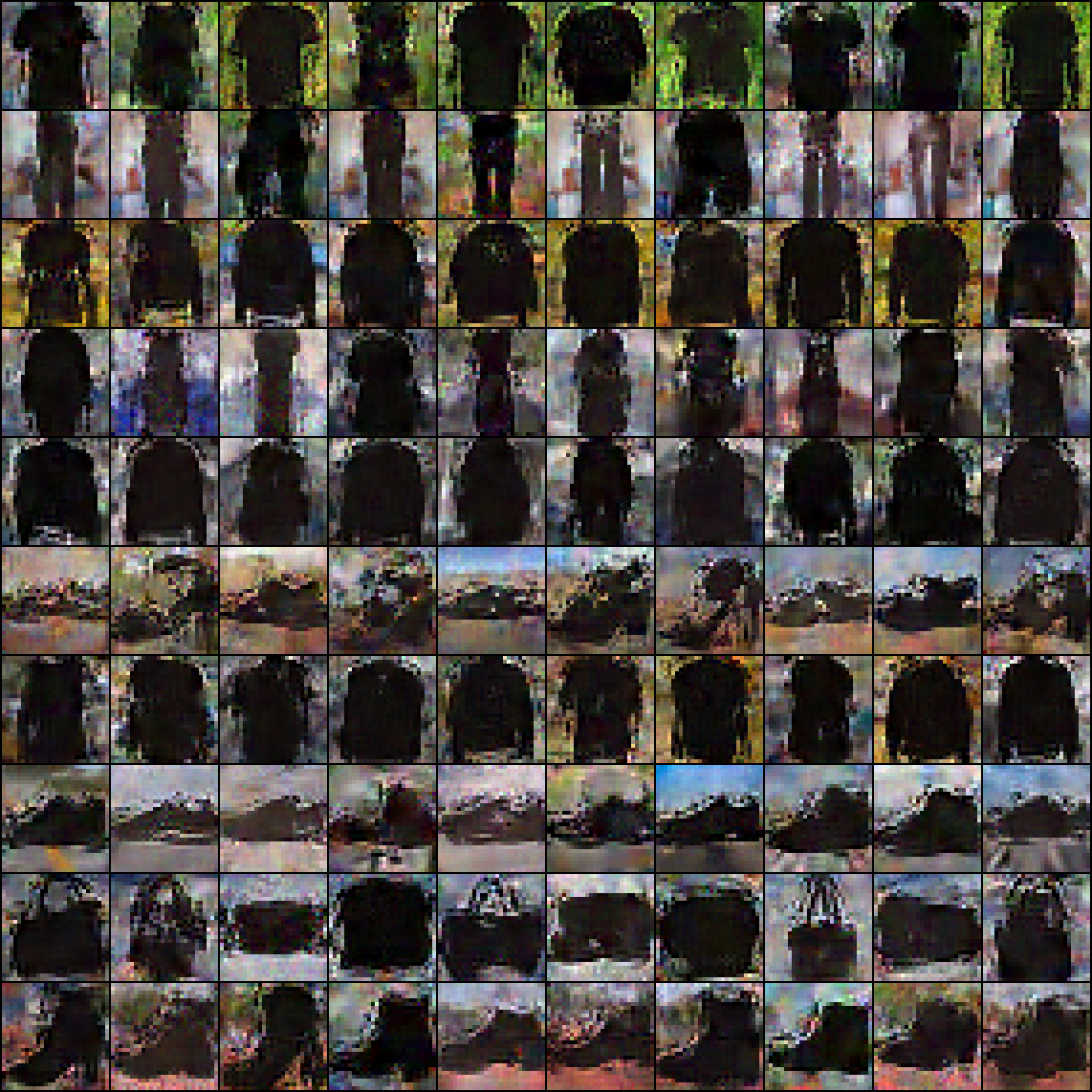}
    \caption{Synthesized dataset by DSA on BG FMNIST with 5\% bias-conflicting samples.}
\end{figure*}

\begin{figure*}
    \centering
    \includegraphics[width=\textwidth]{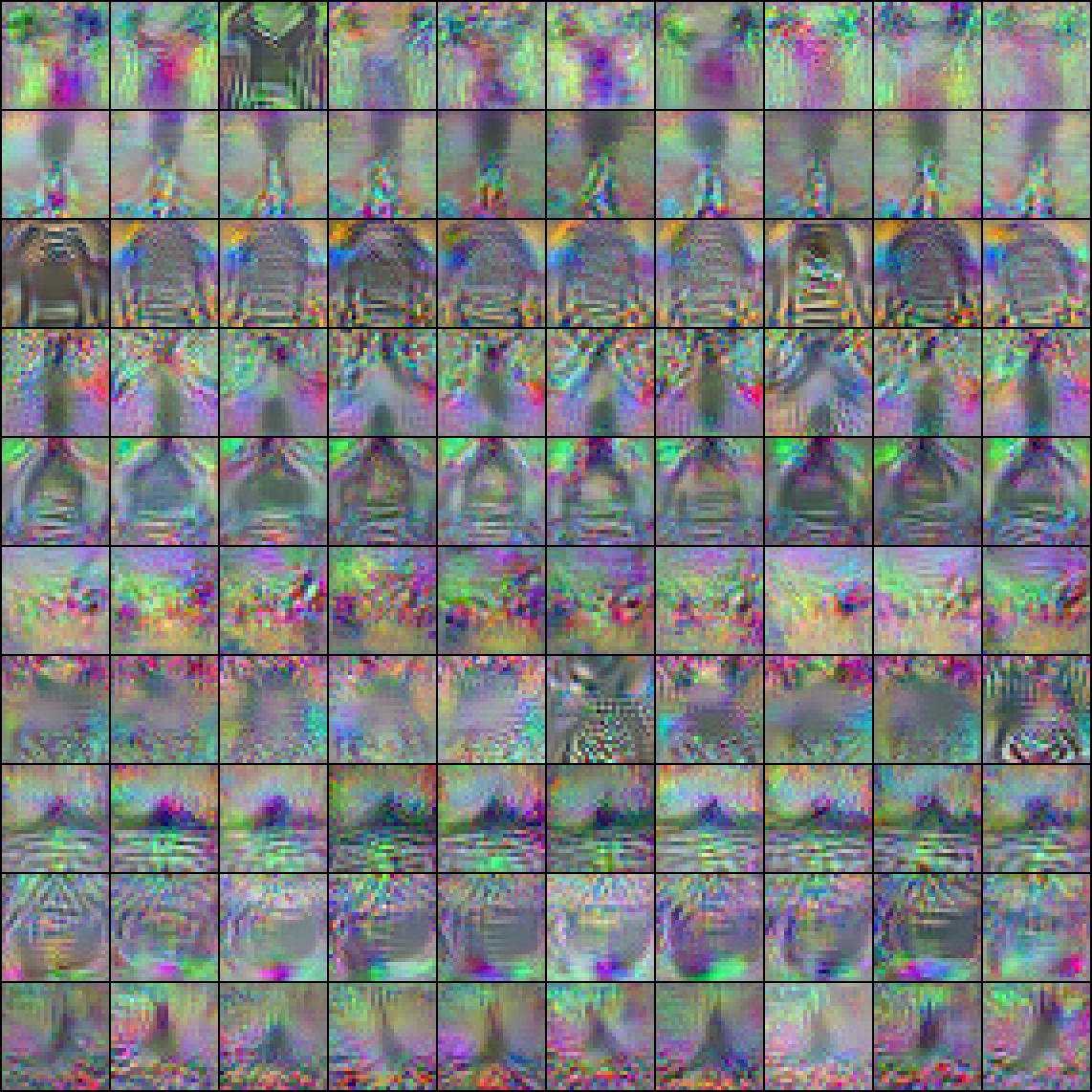}
    \caption{Synthesized dataset by MTT on BG FMNIST with 5\% bias-conflicting samples.}
\end{figure*}

\begin{figure*}
    \centering
    \includegraphics[width=\textwidth]{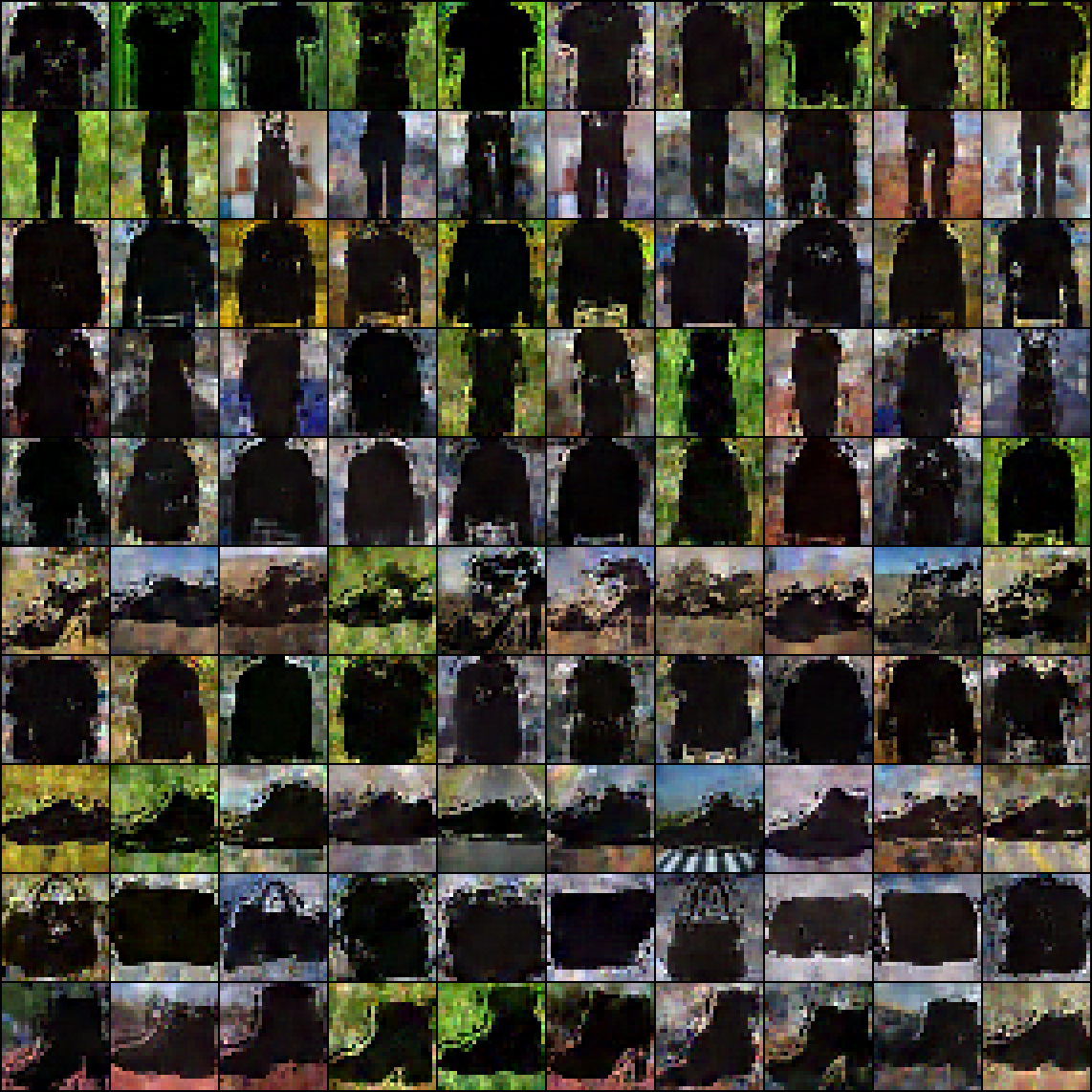}
    \caption{Synthesized dataset by DM+Ours on BG FMNIST with 5\% bias-conflicting samples.}
\end{figure*}

\begin{figure*}
    \centering
    \includegraphics[width=\textwidth]{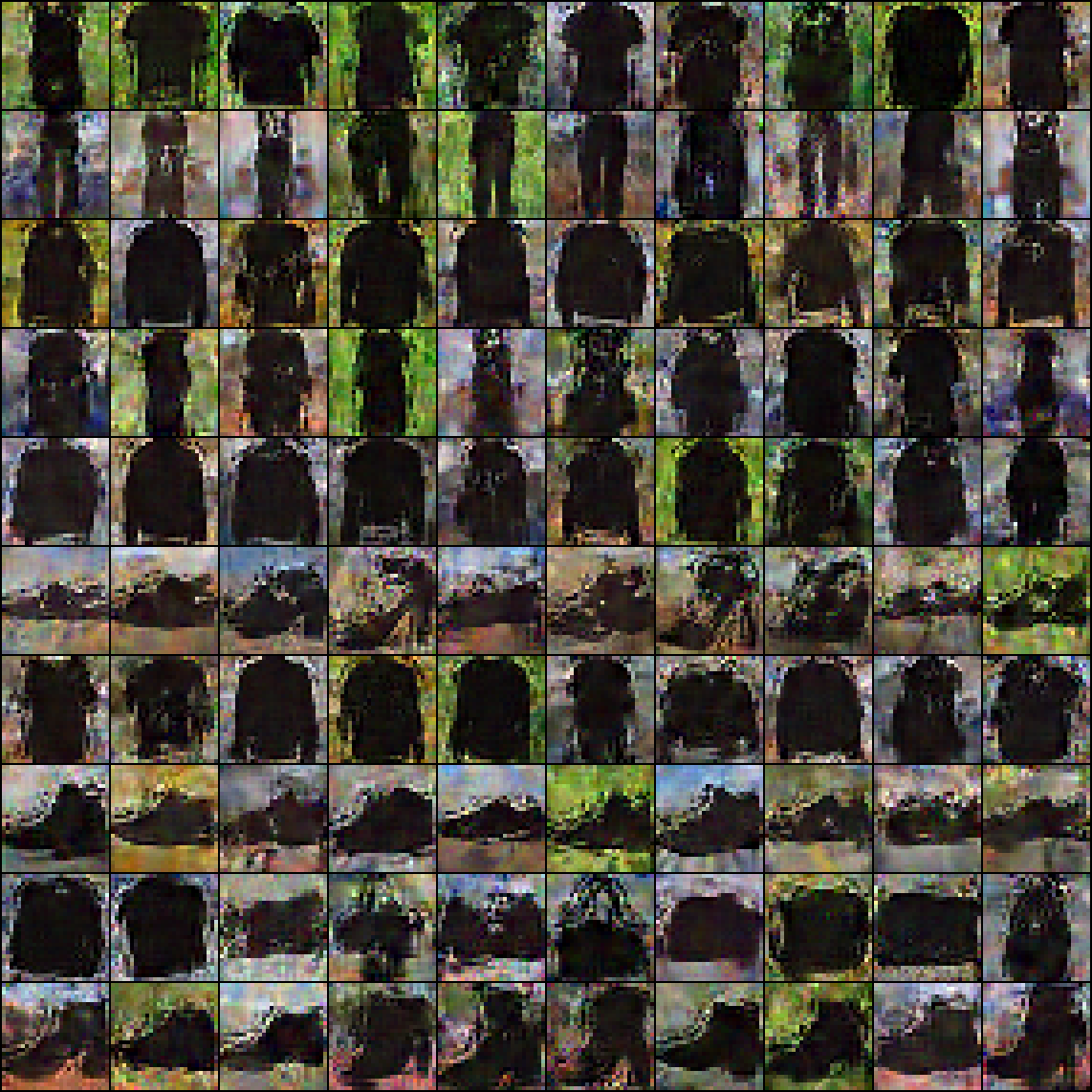}
    \caption{Synthesized dataset by DSA+Ours on BG FMNIST with 5\% bias-conflicting samples.}
\end{figure*}

\end{document}